\PassOptionsToPackage{table,xcdraw,dvipsnames}{xcolor}
\documentclass[acmsmall,nonacm,screen,11pt]{acmart}
\setcopyright{none}
\acmYear{2026}
\usepackage{brand}
\usepackage{header}
\usepackage{docspec}
\renewfontfamily\AppliedDisplayBold{AppliedSansDisplay-Bold}[
  Path=fonts/, Extension=.ttf, BoldFont=AppliedSansDisplay-Bold]

\usepackage{nicefrac}
\usepackage{hyphenat}
\usepackage{algorithm,algpseudocode}

\newcommand{\ourmethod}{ReMind}

\renewcommand{\paperurl}{https://remind-applied.github.io/}

\makeatletter
\patchcmd{\@mkauthors}{\vspace{1.5em}}{\vspace{0pt}}{}{%
  \PackageError{report}{Could not adjust author spacing}{Check the local brand.sty author hook.}}
\patchcmd{\@mkteasers}{\par\bigskip\bgroup}{\par\smallskip\bgroup}{}{%
  \PackageError{report}{Could not adjust teaser spacing}{Check the local acmart teaser hook.}}
\patchcmd{\@mkteasers}{\medskip}{\smallskip}{}{%
  \PackageError{report}{Could not adjust abstract spacing}{Check the local acmart teaser hook.}}
\makeatother

\begin{document}
\title[ReMind: Dynamic Memory for Video Generation]{Teaching Video Generators to Remember:\\Eliciting Dynamic Memory for\\Out-of-Sight State Evolution}

\author{Tianshuo Xu}
\affiliation{\institution{Applied Intuition}\city{Mountain View}\state{CA}\country{USA}}
\author{Yichen Xie}
\affiliation{\institution{Applied Intuition}\city{Mountain View}\state{CA}\country{USA}}
\affiliation{\institution{University of California, Berkeley}\city{Berkeley}\state{CA}\country{USA}}
\author{Depu Meng}
\affiliation{\institution{Applied Intuition}\city{Mountain View}\state{CA}\country{USA}}
\author{Chensheng Peng}
\affiliation{\institution{Applied Intuition}\city{Mountain View}\state{CA}\country{USA}}
\affiliation{\institution{University of California, Berkeley}\city{Berkeley}\state{CA}\country{USA}}
\author{Quentin Herau}
\affiliation{\institution{Applied Intuition}\city{Mountain View}\state{CA}\country{USA}}
\author{Bo Jiang}
\affiliation{\institution{Applied Intuition}\city{Mountain View}\state{CA}\country{USA}}
\author{Yihan Hu}
\affiliation{\institution{Applied Intuition}\city{Mountain View}\state{CA}\country{USA}}
\author{Wei Zhan}
\affiliation{\institution{Applied Intuition}\city{Mountain View}\state{CA}\country{USA}}
\affiliation{\institution{University of California, Berkeley}\city{Berkeley}\state{CA}\country{USA}}

\compactauthors{%
  {\raggedright
   {\rmfamily\bfseries
    \mbox{Tianshuo Xu\textsuperscript{1}}, \;
    \mbox{Yichen Xie\textsuperscript{1,2}}, \;
    \mbox{Depu Meng\textsuperscript{1,\S}}, \;
    \mbox{Chensheng Peng\textsuperscript{1,2}},\\[0.25em]
    \mbox{Quentin Herau\textsuperscript{1}}, \;
    \mbox{Bo Jiang\textsuperscript{1}}, \;
    \mbox{Yihan Hu\textsuperscript{1}}, \;
    \mbox{Wei Zhan\textsuperscript{1,2,\dagger}}\par}
   \smallskip
   {\small\rmfamily\color{AppliedGray}%
    \textsuperscript{1}Applied Intuition \quad
    \textsuperscript{2}University of California, Berkeley\par}
   \smallskip
   {\footnotesize\rmfamily\color{AppliedGray}%
    \textsuperscript{\S}Project lead. \quad \textsuperscript{\dagger}Corresponding author: \texttt{wei.zhan@applied.co}.\par}}
}

\begin{teaserfigure}
\begingroup
\renewenvironment{figure}[1][]{}{}
\begin{figure}[h]
  \centering
  \includegraphics[width=\textwidth]{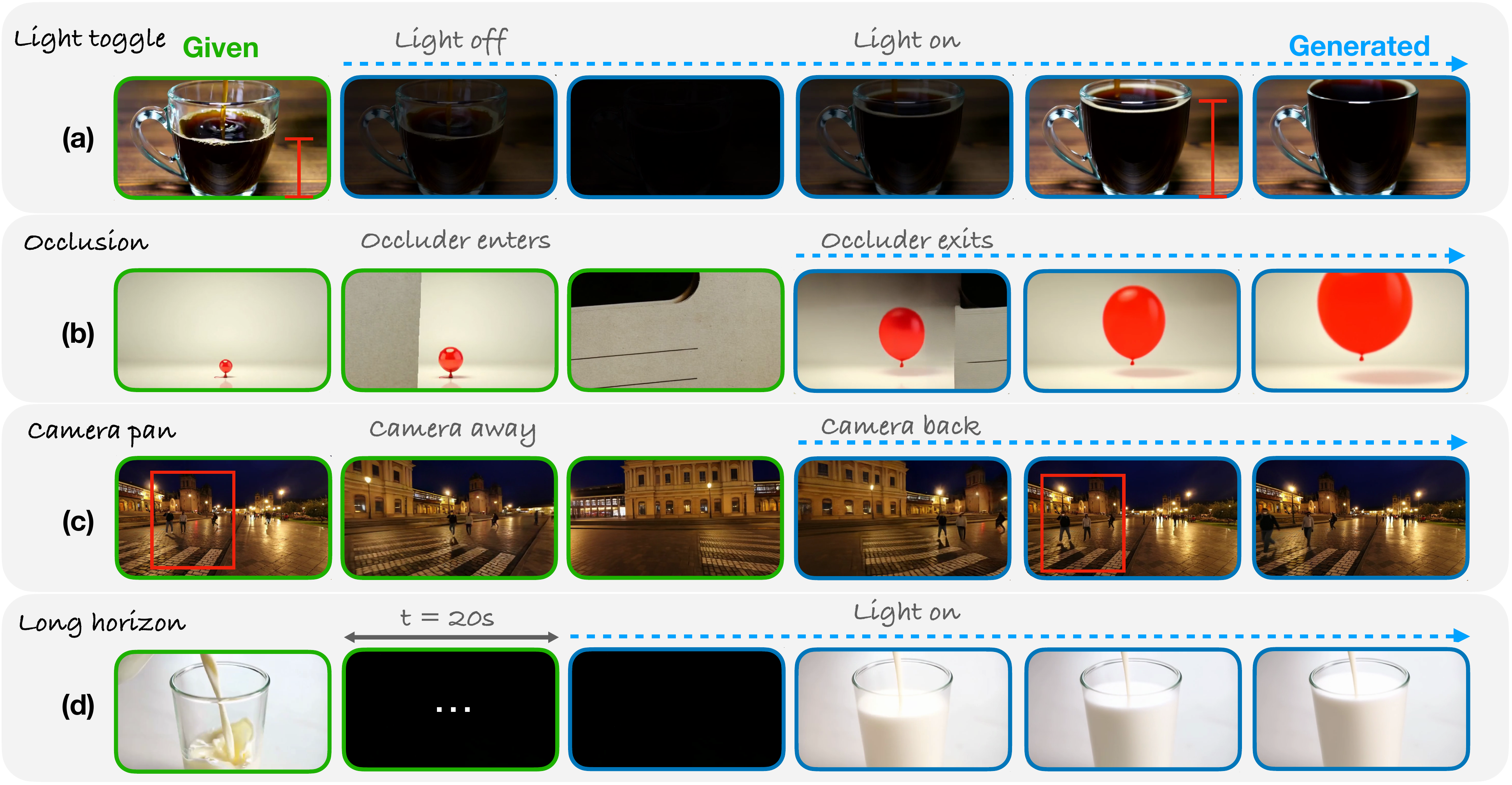}
  \vspace{-1em}
  \caption{Selected qualitative out-of-sight recovery cases. \textbf{(a)} An I2V pouring task with a light-toggle interruption. \textbf{(b)} A V2V balloon-inflation process behind a full-frame occluder. \textbf{(c)} A pedestrian case with camera return. \textbf{(d)} An illustrative 20s dark-interval stress case. These selected examples show visual recovery behavior but do not by themselves establish duration-conditioned state advancement.}
  \label{fig:teaser}
\end{figure}

\endgroup
\end{teaserfigure}

\begin{abstract}

Video world models should maintain evolving states when evidence is unobserved, yet current generators often freeze hidden states upon interruption. This is not simply a capacity problem: pretrained video diffusion transformers already possess KV-cache mechanisms capable of non-local retrieval, but they are rarely trained to use them as dynamic memory. We introduce \ourmethod{}, a framework eliciting dynamic memory behavior via memory-oriented data, event-aware training, and cache adaptation. Organized around a taxonomy of 100+ dynamic events, we build a camera-annotated training mixture combining VLM-filtered real videos, generated hard dynamics, synthetic camera loops, and memory-interruption augmentations. Each clip is converted into a frame graph with protected anchors, degraded intervals, and explicit temporal gaps. A node-structured curriculum---including node-drop, noisy memory, frontier continuation, and reference-cache training---forces the model to retrieve relevant past states across interruptions rather than relying solely on local continuity. PM-RoPE, a camera-phase RoPE extension, provides spatiotemporal addressing through one attention path while preserving pretrained pathways. \ourmethod{} achieves the best overall STEVO-Bench score and improves controlled short-horizon recovery. 


\end{abstract}

\maketitle
\hypersetup{pdftitle={Teaching Video Generators to Remember: Eliciting Dynamic Memory for Out-of-Sight State Evolution},pdfauthor={Tianshuo Xu, Yichen Xie, Depu Meng, Chensheng Peng, Quentin Herau, Bo Jiang, Yihan Hu, Wei Zhan}}
\clearpage
\section{Introduction}

Physical state continues to evolve even when it is temporarily unobserved. A container should keep filling while the lights are off, a balloon should continue inflating behind an occluder, and a pedestrian should keep walking while the camera looks away. Yet current video world models often tie state evolution too closely to immediate visual evidence: when the scene is revealed again, the hidden state may freeze, reset, or resume from an implausible point. This out-of-sight failure, recently formalized in STEVO-Bench~\citep{ma2026outofsight}, exposes a gap between generating visually plausible local motion and maintaining persistent world state across interrupted observation.

At first glance, these failures might seem addressable by scaling temporal
context or improving autoregressive rollout. Self Forcing and LongLive use
self-generated history, streaming tuning, and KV caching for causal
generation~\citep{huang2025selfforcing,yang2025longlive}; Helios instead uses
compressed history injection without a KV cache~\citep{yuan2026helios}. Cache
compression methods further improve storage efficiency~\citep{ranganath2026kv}.
In parallel, memory-conditioned generators inject historical views or
3D/4D representations to improve spatial consistency under camera
motion~\citep{yu2025cam,xiao2025worldmem,ren2025gen3c,yang2026neoverse,inspatio2026world}.
However, these advances do not directly teach a model when an older observation
should override recent but corrupted context. We view the missing behavior
through the lens of frame graphs: standard training presents frames as a local
chain, while interruptions create non-local edges from a recovery frame to the
latest reliable state anchor. If training never exposes such edges, attention
and KV-cache machinery may store past tokens without learning when to retrieve
them as dynamic memory.

We introduce \ourmethod{}, a framework for eliciting dynamic memory behavior from a pretrained autoregressive video diffusion transformer~\citep{wan2025}. \ourmethod{} combines memory-oriented data construction, event-aware training, and pretraining-compatible cache adaptation. We build a camera-annotated training mixture organized around a taxonomy of over 100 dynamic events, combining VLM-filtered real videos, generated hard dynamics, rendered camera loops, and memory-interruption augmentations. Each clip is converted into a frame graph with protected state anchors, degraded observation intervals, and explicit temporal gaps, defining which historical nodes should remain useful when local visual evidence becomes unreliable.

Given these memory graphs, \ourmethod{} trains the model with a node-structured curriculum, including node-drop, noisy memory, V2V frontier, and reference-cache training. These regimes corrupt, drop, or temporally separate observations so that successful recovery requires retrieving the relevant past state rather than relying only on local continuity. To make such retrieval geometrically meaningful, we add Projective Memory RoPE (PM-RoPE), a camera-phase extension to rotary position embedding (RoPE)~\citep{su2024roformer}. Unlike dual-attention mechanisms adding a separate spatial branch, PM-RoPE grants cached entries unified spatiotemporal addresses within a single self-attention operation, preserving their original temporal positions and the pretrained attention pathway.

We evaluate \ourmethod{} on STEVO-Bench~\citep{ma2026outofsight} and controlled recovery tasks covering occlusion, darkness, and camera lookaway. \ourmethod{} achieves the best overall score among compared models, with leading state-progress and competitive physical-plausibility and coherence scores. Matched 5B VBench arms have similar point estimates, without an equivalence or non-inferiority claim. We additionally use KV-importance maps as qualitative diagnostics of whether recovery is associated with attention to intended historical anchors.

Our main contributions are summarized as follows:
\begin{itemize}
  \item We formulate out-of-sight state evolution as a dynamic-memory problem over frame graphs and introduce \ourmethod{}, a memory-elicitation framework constructing camera-annotated training graphs with protected anchors, degraded intervals, and temporal gaps.
  \item We propose a node-structured training framework with pretraining-compatible cache adaptation: node-drop, noisy memory, V2V frontier, and reference-cache training force retrieval of reliable state anchors, while PM-RoPE provides unified spatiotemporal addressing through one attention path when paired with non-contiguous cache positioning.
  \item We establish a validation protocol that measures both hidden-state recovery and standard generation quality, combining STEVO-Bench evaluation, controlled V2V recovery cases, reference-cache stress tests, KV-importance diagnostics, and VBench evaluation.
\end{itemize}

\section{Related Work}

\begin{figure}[t]
    \centering
    \includegraphics[width=\linewidth]{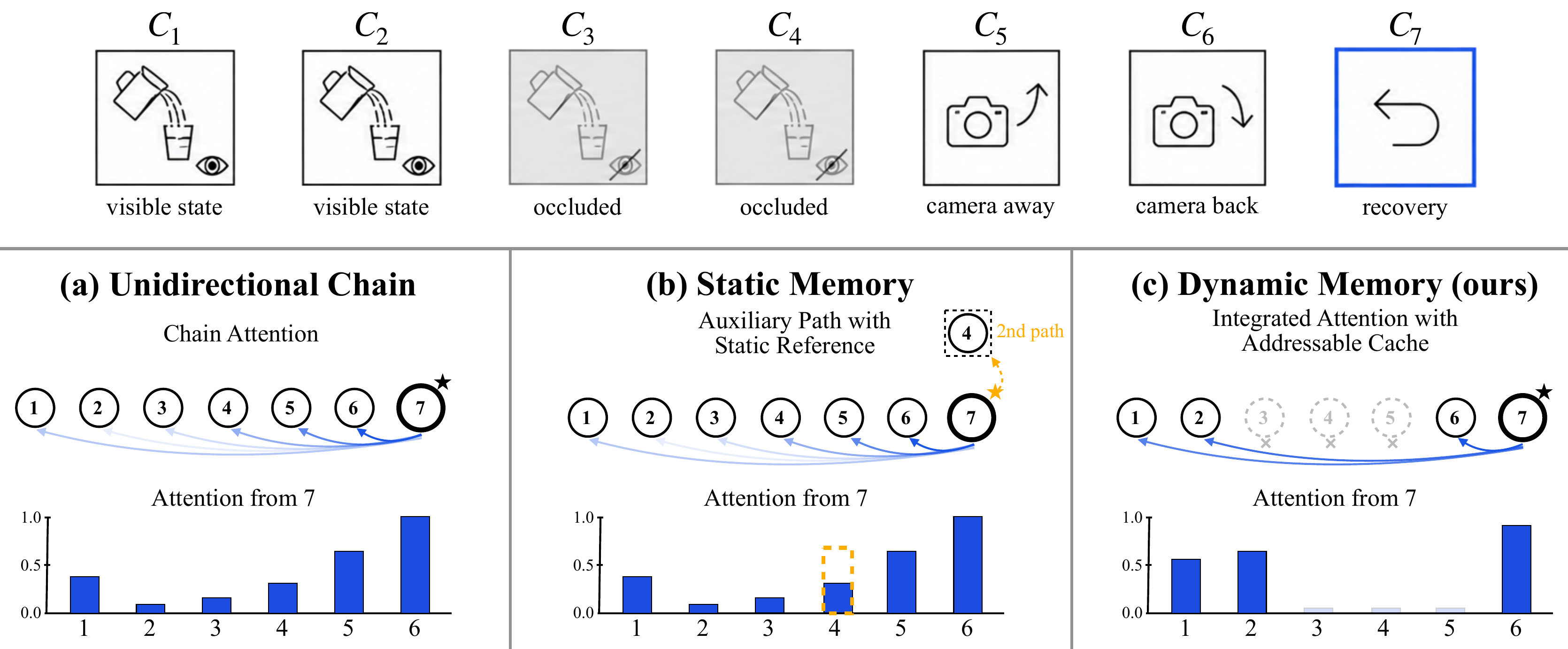}
    \caption{Conceptual comparison of memory use in video generation.
    \textbf{(a) Unidirectional Chain:} KV-cached models primarily retain recent
    rollout context. \textbf{(b) Structured Spatial Memory:} memory- or
    world-conditioned methods retrieve geometry-aligned historical references
    for spatial and revisit consistency. \textbf{(c) \ourmethod{} (Dynamic
    Memory):} our training exposes the native cache to reliable anchors,
    degraded observations, and evolving targets so it can use non-local
    historical evidence for state recovery.}
    \label{fig:graph}
\end{figure}

\subsection{Autoregressive Diffusion Video Generation}
Autoregressive diffusion adapts pretrained video diffusion backbones such as
Wan~\citep{wan2025} into causal rollout models. Teacher forcing conditions on
clean history~\citep{williams1989learning}, Diffusion Forcing uses independent
per-token noise levels~\citep{chen2024diffusion}, and Self Forcing reduces the
train-test mismatch by rolling out generated context with KV
caching~\citep{huang2025selfforcing}. LongLive uses streaming-long tuning,
short-window attention, and KV recaching~\citep{yang2025longlive}, whereas
Helios uses compressed history injection without a KV cache or causal
masking~\citep{yuan2026helios}. These systems primarily target long-video
quality or throughput rather than selecting an older reliable state witness.
Separate cache compression and pruning studies improve
efficiency~\citep{ranganath2026kv}, but do not teach the model when to retrieve
an older observation as the latest reliable state witness. \ourmethod{} targets this out-of-sight failure~\citep{ma2026outofsight}
by training existing attention and cache mechanisms to behave as dynamic memory.



\subsection{Memory Mechanisms for Video World Models}

There are already some explorations regarding the memory mechanism for video generation models. We conceptually categorize them by their underlying topologies in Fig.~\ref{fig:graph}.

\textbf{Unidirectional Chain:} KV-cached causal models commonly retain recent
rollout context as a temporal buffer~\citep{huang2025selfforcing,yang2025longlive}
rather than an explicitly indexed scene-state store, which may leave them
vulnerable when recent observations are corrupted.

\textbf{Structured Spatial Memory:} Recent methods inject spatial or 3D
representations into memory. WorldPlay~\citep{sun2025worldplay} dynamically
reconstitutes a context set and uses an additional PRoPE attention path
alongside standard temporal RoPE~\citep{su2024roformer,li2025prope}.
InSpatio-World~\citep{inspatio2026world} and NeoVerse~\citep{yang2026neoverse}
use explicit geometric constraints or rendered projections, while
MosaicMem~\citep{yu2026mosaicmem} retrieves spatially aligned 3D patches.
These methods primarily target geometric or revisit consistency; our focus is
how an evolving state advances while it is unobserved.

\textbf{Dynamic Memory:} Our method, \ourmethod{}, moves beyond static memory by treating the KV cache in a dynamic way. Inter-frame relevance is not fixed to pure temporal adjacency or spatial overlapping. Instead, the model learns to adaptively condition the generation on past frames through its own attention module with our proposed PM-RoPE and training regimes, which forces the model to resort to most related historical anchors under various distractors.

\subsection{Camera Control for Video Generation}

Camera-controlled video generation injects pose information through camera
encoders, ControlNet-style modules, geometry-aware conditions, or attention
modifications~\citep{he2024cameractrl,bahmani2024vd3d,ren2025gen3c,yang2026neoverse,inspatio2026world}.
Recent attention-native methods further encode camera geometry through
projection-aware transports, parallel camera branches, ray-coordinate RoPE, or
relative camera pose embeddings~\citep{li2025prope,sun2025worldplay,zhang2025ucpe,xie2026raynova,li2026rerope,xie2026urope}.
Motion Forcing~\citep{xu2026motionforcing} and STANCE~\citep{stance2026}
instead target motion and interaction control rather than camera-trajectory
control.
PM-RoPE serves a narrower role in \ourmethod{}: it provides camera-aware
addresses and non-contiguous temporal positions for cached anchors while
preserving the pretrained attention pathway, enabling the event-aware
curriculum to train non-local state retrieval.


\section{\ourmethod{} with Dynamic Memory}
\label{sec:method}

\begin{figure}[t]
    \centering
    \includegraphics[width=\linewidth]{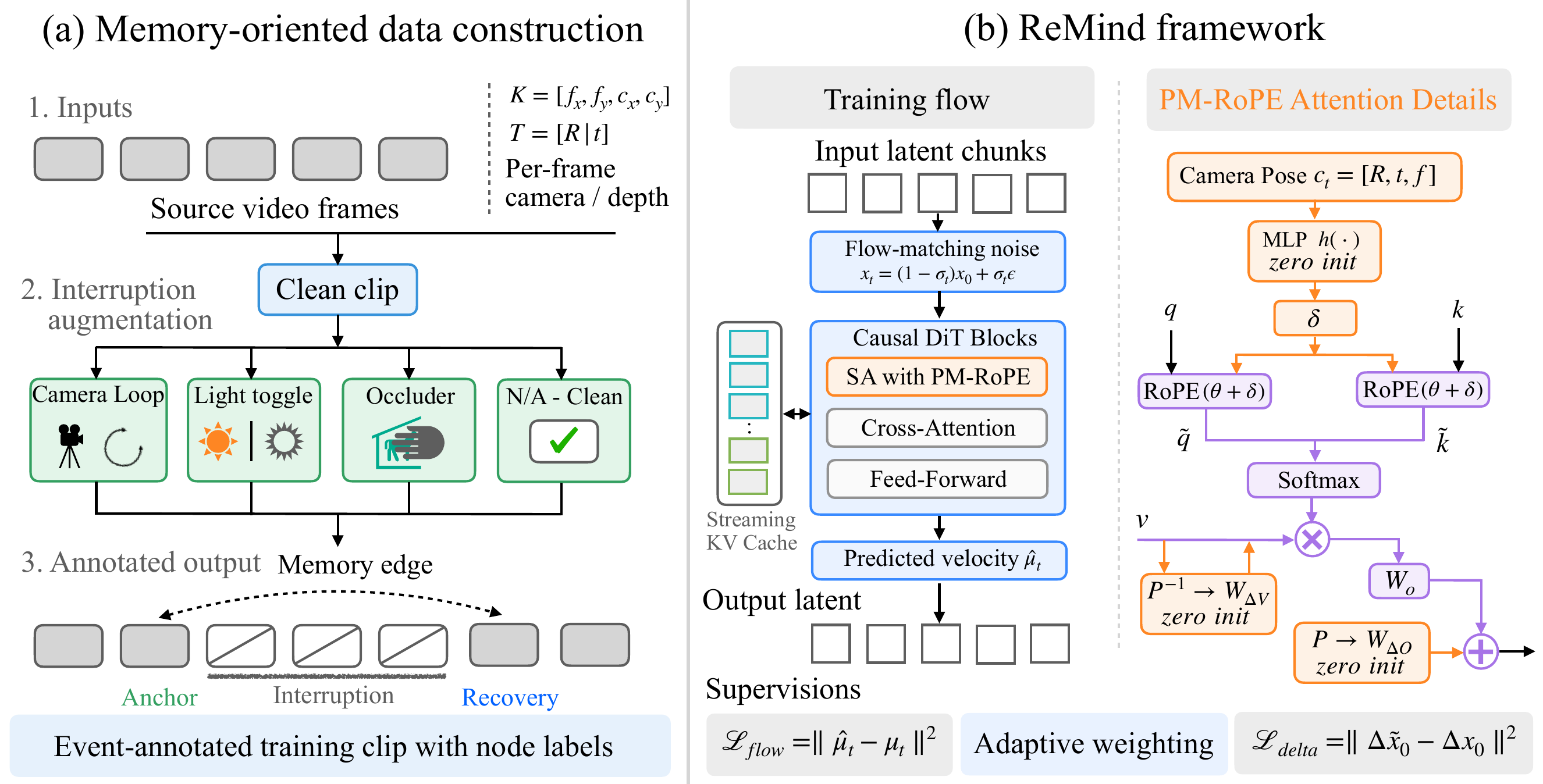}
    \caption{\textbf{Overview of \ourmethod{}.} \textbf{(a)} Four parallel data
    streams---general videos, natural dynamics, generated hard dynamics, and
    rendered camera loops---are normalized independently. Interruption
    augmentations are then applied within a clip; no cross-clip concatenation or
    insertion into DL3DV is performed. The resulting frame graphs contain
    annotated anchor, interruption, and recovery nodes. \textbf{(b)} The causal
    video DiT uses its native KV cache as dynamic memory. Cached chunks retain
    non-contiguous RoPE positions, while PM-RoPE adds camera-conditioned phase
    offsets and zero-initialized value/output residuals.}
    \label{fig:arch}
\end{figure}

\ourmethod{} builds on a pretrained causal video diffusion transformer and
elicits dynamic memory through memory-oriented data, cache-compatible spatial
addressing, and node-structured training. Each chunk can
attend to cached entries that retain their original temporal positions and
camera-aware addresses, allowing the model to retrieve historical state anchors
rather than only recent context. Sec.~\ref{sec:dataset} describes the
memory-oriented data construction, Secs.~\ref{sec:projective-memory-rope}
and~\ref{sec:kv-memory} introduce PM-RoPE and the non-contiguous streaming KV
cache, and Sec.~\ref{sec:memory-training} presents the training curriculum that
teaches the model to use this memory adaptively. The overall framework of \ourmethod{} is illustrated in Fig.~\ref{fig:arch}.

\subsection{Memory-Oriented Data Construction}
\label{sec:dataset}

We construct our training data to strengthen the dynamic memory of \ourmethod{} from public video sources, OpenVid~\citep{nan2024openvid} and DL3DV~\citep{ling2023dl3dv}.  Since these sources are not targeted at dynamic-memory
events, we augment them with a taxonomy of over one hundred dynamic event types,
scored by visual salience, frequency, and usefulness for hidden-state recovery.
The taxonomy drives Pexels retrieval with dynamics-specific keywords and
Qwen3-VL filtering~\citep{bai2025qwen3vl}, where the VLM captions events,
detects cuts, and estimates the spatial and temporal occupancy of the dynamic
region.  We further synthesize underrepresented high-occupancy dynamics, such
as close-up pouring/filling, and render selected dynamic scenes under loop
trajectories with Neoverse~\citep{yang2026neoverse} for camera-memory
supervision. The full dynamic taxonomy and data
collection policy are provided in Appendix~\ref{app:data}.

These sources are four parallel streams, not stages of a single synthesis
pipeline: general OpenVid/DL3DV clips are used without injected objects;
natural Pexels dynamics remain real clips; hard dynamics are generated
independently; and NeoVerse is used only for rendered camera-loop supervision.
Orthogonally, light, occluder, zoom, and camera perturbations are applied
within one clip. We never concatenate unrelated clips or insert generated
content into DL3DV.


We then convert clips into memory-training graphs by applying interruption
augmentations inspired by STEVO-Bench~\citep{ma2026outofsight}, including
camera loops, light toggles, moving occluders, and zoom/camera perturbations.
All clips are normalized with RGB frames, captions, event metadata, and, when
available, camera intrinsics/extrinsics and depth; for web videos without
native geometry, we estimate depth and camera trajectories with Depth Anything
3~\citep{lin2025depth}.  Event nodes are derived from loop-return points or
known peak/recovery frames, and define the protected anchors and recovery
regions used by the memory curriculum.

\subsection{Projective Memory Rotary Positional Embedding}
\label{sec:projective-memory-rope}

\textbf{Motivation: why couple camera and temporal addressing?}
PM-RoPE is an effective inductive bias, not a theoretical expressivity
necessity: sufficiently deep decoupled spatial and temporal attention can mix
both signals. However, WorldPlay-style designs~\citep{sun2025worldplay}
instantiate a second attention path and do not expose the conjunction of
camera address and pre-degradation time in a single score. Under same-position
occlusion, the spatial term ties the clean anchor and recent occluder, leaving
finite finetuning to infer the boundary through temporal cache order and
ambiguous content (Appendix~\ref{app:math-proof}). We therefore introduce
PM-RoPE, a camera-conditioned extension of rotary position embedding
(RoPE)~\citep{su2024roformer}. It folds camera-dependent phase offsets and
zero-initialized value/output residuals into the pretrained self-attention path,
providing an explicit joint spatiotemporal address without a second attention
operation while preserving the backbone Q/K/V/O projections. The computational
advantage is empirical rather than assumed; Table~\ref{tab:efficiency} reports
the measured latency, memory, cache, and FLOPs.

For each frame $i$, we build a compact pose descriptor $\mathbf{c}_i$ based on the current camera extrinsic parameters $\mathbf{P}_i=\left[\mathbf{R}_i|\mathbf{t}_i\right]$ and focal lengths $(f^x_i,f^y_i)$ as:
\begin{equation}
\mathbf{c}_i = \left[\operatorname{vec}(\mathbf{R}_i),\mathbf{t}_i,\log f^x_i,\log f^y_i\right],
\end{equation}
where translation and intrinsics are normalized. The released implementation is
\emph{focal-only}: it retains normalized $(f_x,f_y)$ and sets the principal-point
offset $(c_x,c_y)$ to zero rather than consuming the full intrinsic matrix. In
each self-attention layer, a zero-initialized MLP $h(\cdot)$ maps $\mathbf{c}_i$
to a per-frame rotary phase offset $\delta_i$, which is added to the standard
spatiotemporal phase in RoPE to indicate the relative positions between the
camera poses at different frames:
\begin{equation}
\begin{aligned}
\tilde q_i &= \operatorname{RoPE}_{\theta_i+\delta_i}(q_i),\qquad \delta_i = h(\mathbf{c}_i) \\
\tilde k_j &= \operatorname{RoPE}_{\theta_j+\delta_j}(k_j),\qquad \delta_j = h(\mathbf{c}_j)
\label{eq:pm_rope}
\end{aligned}
\end{equation}

Inspired by PRoPE~\citep{li2025prope}, we inject camera-dependent transforms into the value and output paths of the self-attention module with zero-initialized residual projections $W_{\Delta V}$ and $W_{\Delta O}$:
\begin{equation}
\begin{aligned}
\tilde v_j = v_j + W_{\Delta V}(\mathcal{P}_j^{-1}v_j),\qquad
\tilde y_i =
\sum_{j \leq i}
\operatorname{softmax}_j
\left(
\frac{\tilde q_i^\top \tilde k_j}{\sqrt{d}}
\right)
\tilde v_j,\qquad
o_i = W_O \tilde y_i + W_{\Delta O}(\mathcal{P}_i\tilde y_i).
\end{aligned}
\label{eq:projective-memory-rope}
\end{equation}
where $\mathcal{P}_j^{-1} v_j$ denotes the feature-map $v_j$ back-projected from the camera coordinate system of frame $j$ into the world coordinate system (and $\mathcal{P}_i \tilde y_i$ represents the subsequent projection into the query frame $i$). In our implementation, this is achieved by linear projections $W_{\Delta V}, W_{\Delta O}$ operating on the concatenated latent and normalized 6-DoF camera embeddings.

Since $W_{\Delta V}$, $W_{\Delta O}$, and $h(\cdot)$ are initialized as zeros, in the beginning of our finetuning process, PM-RoPE is the same as RoPE in the pretraining with $\delta_i=0$ and $\tilde{v}_j=v_j$.
As the finetuning process progresses, the PM-RoPE gradually takes the camera information into consideration so that the auto-regressive video generation can adaptively exploit historical frames according to the inter-frame relative camera poses in an adaptive manner.

\subsection{Dynamic Memory with Streaming KV Cache}
\label{sec:kv-memory}

We partition the latent video into chunks and train with chunk-causal attention. During inference, dynamic memory is implemented as a streaming KV cache for auto-regressive video generation. Each generated chunk writes its keys and values into a KV cache for later chunks to read.

To ensure temporal plausibility, the key design is to preserve the original frame position of each cached chunk instead of treating them equivalently or compacting all cached entries into consecutive timesteps. This prevents old references from being confused with recent neighbors. For
example, a clean reference chunk may be placed at positions $0,\ldots,m-1$,
while the target video starts after a sampled gap of $G$ chunks, at
$Gm,\ldots,Gm+T-1$. Thus, the cache can store sparse, non-contiguous memories while still telling the model how far each memory item is from the current state.

\subsection{Training Scheme for Dynamic Memory}
\label{sec:memory-training}

To encourage the model to exploit the dynamic memory, we propose a novel strategy.
We train with a flow-matching objective.
For clean latent $x_0$, Gaussian noise $\epsilon$, and scheduler value
$\sigma_t$, the noised input and target are:
\begin{equation}
x_t=(1-\sigma_t)x_0+\sigma_t\epsilon,\qquad
u_t=\epsilon-x_0.
\end{equation}
The model predicts $u_t$ and is optimized with a masked mean-squared loss
over selected chunks. The training process includes multiple schemes, decided by different masks and timestep schedules:

The curriculum contains four complementary regimes. \emph{All-history training}
uses a chunk-causal pass in which each chunk attends to all previous chunks,
matching the basic streaming inference path. \emph{Noisy-memory} and
\emph{node-drop} training corrupt past chunks with high noise timesteps or
replace interruption nodes with pure-noise latents while preserving at least one
event anchor, forcing the model to ignore unreliable recent context and recover
from surviving historical evidence. \emph{V2V suffix/frontier training} keeps a
clean or degraded prefix and supervises only the recovery suffix, matching
chunk-wise autoregressive deployment. \emph{Reference-cache training} prepends
clean reference chunks from the undegraded clip at old positions and starts the
target video after a sampled temporal gap, directly teaching retrieval from
non-contiguous memory. Together, these regimes train the model to compare memory
candidates with different visual reliability, camera poses, RoPE positions, and
state histories, rather than attending only to a continuous prefix or fixed
reference frame. Algorithm~\ref{alg:training-step} gives the complete sampling,
corruption, masking, and optimization procedure for one training step.

\noindent\textbf{Dynamic Auxiliary Loss and Adaptive Weighting.}
Localized dynamic events often occupy only a small fraction of the canvas, so a standard flow-matching loss can be dominated by static background reconstruction. We therefore add a dynamic temporal-delta loss with adaptive weighting after warmup, increasing its influence for localized low-variance dynamics and reducing it for large camera or scene changes. Appendix~\ref{app:dynamic-loss} gives the full objective and hyperparameters.


\begin{figure}[t]
    \centering
    \includegraphics[width=\linewidth]{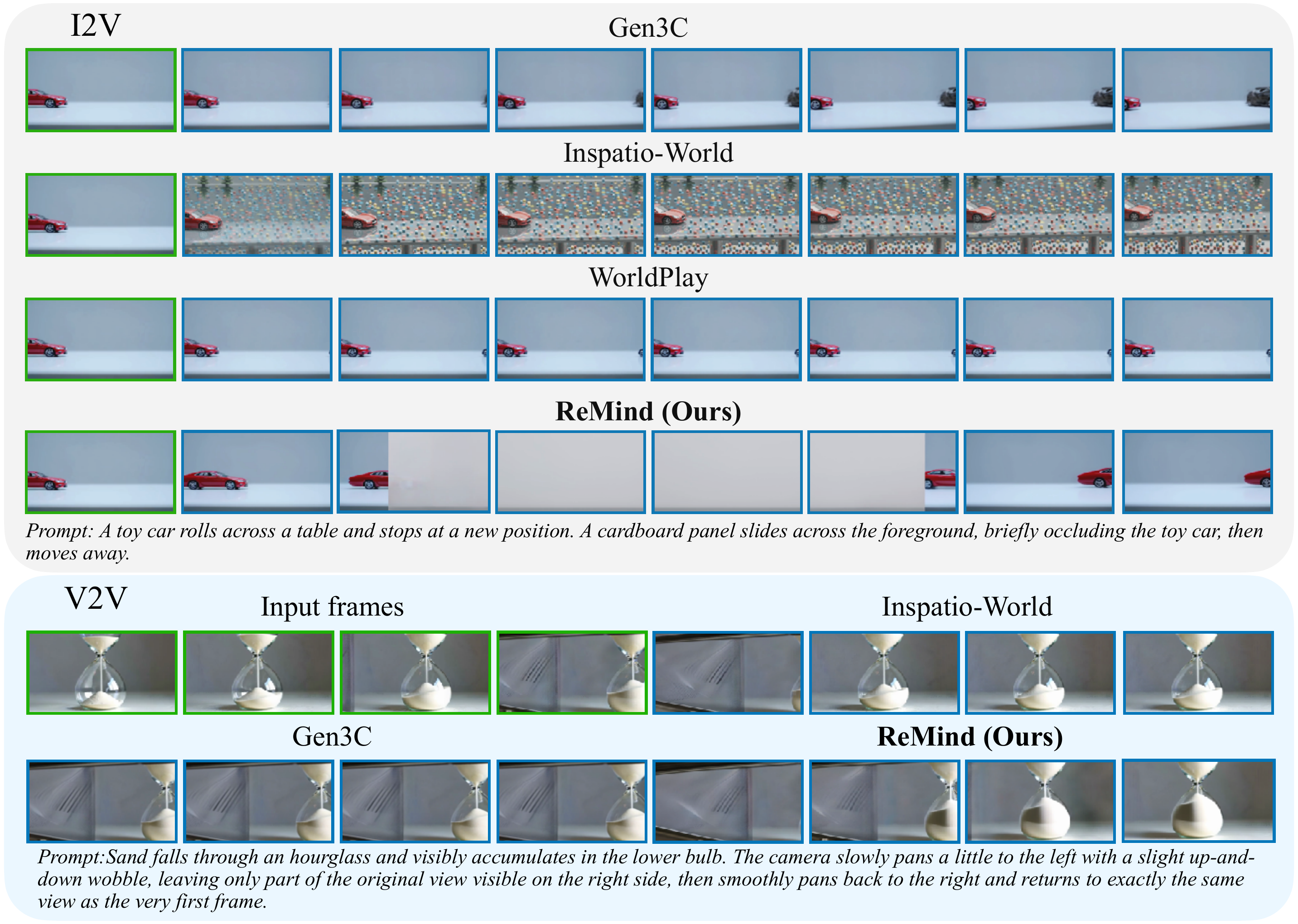}
    \caption{Qualitative comparisons with recent methods. WorldPlay is omitted
    from the controlled V2V recovery panel because no released result or
    interface is available for our degraded-prefix recovery protocol.}
    \label{fig:comparison}
\end{figure}

\section{Experiments}
\label{sec:experiments}


\subsection{Evaluation Suite}
\label{sec:evaluation}

We evaluate \ourmethod{} on both targeted hidden-state recovery and general I2V
quality. For hidden-state recovery, we use the STEVO-Bench I2V
protocol~\citep{ma2026outofsight}, where each example provides an initial image
and an instruction describing an evolving process under occlusion, lights-off,
or camera-lookaway distractors. We report the benchmark's five metrics: \emph{State
Progress} measures whether the hidden physical state evolves correctly; \emph{Physical
Plausibility} measures artifact-free dynamics; \textit{Coherence} measures temporal scene
consistency; \emph{Observation Control} measures whether the requested distractor is
applied; and \emph{Action Control} measures whether the instructed action occurs. For
general generation quality, we use VBench~\citep{huang2024vbench} to assess
semantic accuracy, temporal quality, and visual quality in standard I2V
settings. We separately use controlled V2V recovery, for which a degraded real
prefix is supplied and the clean source suffix provides time-aligned ground
truth. Appendix~\ref{app:additional-results} defines both protocols and reports
the matched controls and causal memory probes.
\subsection{Quantitative Results: Out-of-Sight vs. General Quality}
\label{sec:results}

A critical requirement for dynamic memory frameworks is the ability to acquire strong out-of-sight state evolution capabilities without suffering from catastrophic forgetting of general image-to-video (I2V) priors. To demonstrate this, we conduct evaluations on two distinct fronts: STEVO-Bench~\citep{ma2026outofsight} for targeted state evolution under occlusion/interruption, and VBench~\citep{huang2024vbench} for general generation quality and consistency.

\textbf{State Evolution (STEVO-Bench~\citep{ma2026outofsight}).}
We evaluate our method against the base model and recent memory-augmented baselines on the STEVO-Bench I2V protocol. As shown in Table~\ref{tab:stevo-results}, \ourmethod{}'s dynamic non-local routing allows it to bridge temporal discontinuities, yielding a substantial improvement in state progress and physical coherence compared to other video models.

\begin{table}[t!]
\caption{Quantitative I2V evaluation on STEVO-Bench. Metrics are percentages.
Baseline scores are the official leaderboard values; our row uses the 1.3B main
submission checkpoint. \ourmethod{} achieves the highest state progress and
class-leading coherence among models without camera-guided image warping.}
\centering
\small
\setlength{\tabcolsep}{4pt}
\begin{tabular}{l c c c c c c}
\toprule
\raisebox{0.75em}{Model} & \shortstack{State \\ Progress $\uparrow$} & \shortstack{Physical \\ Plausibility $\uparrow$} & \raisebox{0.75em}{Coherence $\uparrow$} & \shortstack{Observation \\ Control $\uparrow$} & \shortstack{Action \\ Control $\uparrow$} & \raisebox{0.75em}{Total $\uparrow$} \\
\midrule
\multicolumn{7}{c}{\textit{Video Models}} \\
\midrule
HunyuanVideo 1.5 & 4.1 & 42.1 & 59.1 & 31.2 & \textbf{81.0} & 217.5 \\
WAN 2.2 & 7.7 & 52.0 & 58.4 & 46.2 & 76.5 & 240.8 \\
CogVideoX & 1.4 & 68.5 & 67.1 & 22.2 & 75.5 & 234.7 \\
\midrule
\multicolumn{7}{c}{\textit{Camera-Controlled Models}} \\
\midrule
Gen-3C & 0.0 & 30.6 & 82.4 & \textbf{90.6} & 57.6 & 261.2 \\
HY-WorldPlay & 0.0 & 72.2 & 88.2 & 55.4 & 64.9 & 280.7 \\
Genie 3 & 2.9 & 15.2 & 27.3 & 84.7 & 78.6 & 208.7 \\
Lingbot World & 3.4 & 40.7 & 76.3 & 35.6 & 67.8 & 223.8 \\
InSpatio-World & 4.5 & \textbf{79.6} & \textbf{99.6} & 0.0 & 69.7 & 253.4 \\
\midrule
\textbf{\ourmethod{} (Ours)} & \textbf{14.5} & 71.5 & 96.4 & 41.6 & 76.9 & \textbf{300.9} \\
\bottomrule
\end{tabular}

\label{tab:stevo-results}
\end{table}

As shown in Table~\ref{tab:stevo-results}, while \ourmethod{} does not achieve the highest score in every individual metric, it achieves the highest official \textit{Total} score by balancing these competing requirements. For instance, the InSpatio-World~\citep{inspatio2026world} baseline achieves exceptionally high Coherence (99.6\%) but fails Observation Control (0.0\%). This suggests that it often maintains visual consistency by avoiding requested degradations such as occlusions or lights-off, and therefore does not fully face the recovery challenge. In contrast, \ourmethod{} executes these partial observations (41.6\% Observation Control) while maintaining strong Coherence (96.4\%) and the best State Progress (14.5\%). Genie~3 is closed; its row, like the other baseline rows, is an official leaderboard aggregate rather than a local rerun.
\begin{table}[h]
\centering
\begin{minipage}[t]{0.5\linewidth}
\textbf{General Generation Quality (VBench~\citep{huang2024vbench}).}
Table~\ref{tab:nocamera-results} provides a cross-scale reference for standard
I2V quality. Because InSpatio-World uses a 1.3B checkpoint and \ourmethod{} uses
a 5B checkpoint, this comparison is not a component attribution. The matched
5B data-only/full comparison in Appendix~\ref{app:vbench-detailed} provides
matched point estimates, but no prespecified equivalence margin or confidence
interval.
\end{minipage}
\hfill
\begin{minipage}[t]{0.45\linewidth}
\caption{VBench I2V evaluation under a common runner. InSpatio-World uses its
released 1.3B checkpoint; \ourmethod{} uses the 5B full-method checkpoint.}
\centering
\small
\setlength{\tabcolsep}{3pt}
\begin{tabular}{l c c}
\toprule
Metric Group & InSpatio & \textbf{\ourmethod{}} \\
\midrule
Semantic Accuracy $\uparrow$ & 90.29 & \textbf{98.16} \\
Motion \& Temporal $\uparrow$ & 76.11 & \textbf{77.22} \\
Visual Quality $\uparrow$ & 56.83 & \textbf{65.18} \\
\bottomrule
\end{tabular}
\label{tab:nocamera-results}
\end{minipage}
\end{table}
\subsection{Ablation Study}
\label{sec:ablation}

We conduct ablation studies to analyze three major components in \ourmethod{}:
\textbf{1) The PM-RoPE architecture}, which injects camera geometry into attention;
\textbf{2) the dynamic loss design}, which improves memory-aware optimization; and
\textbf{3) the memory-oriented data construction pipeline}, which targets continuous scene evolution and is analyzed in Appendix~\ref{app:ablation-data}.

\noindent\textbf{Failure archetypes.}
The five columns separate two failures that a single quality score would
conflate. InSpatio-World has 99.6 Coherence but 0.0 Observation Control,
consistent with \emph{evading the interruption}. Conversely, Gen-3C executes
the requested observation control (90.6) but has 0.0 State Progress; the
official HY-WorldPlay result likewise has 0.0 State Progress. These are
\emph{restore-but-freeze} failures: the scene remains coherent or the control
is rendered, but the hidden process does not measurably advance. \ourmethod{}
raises State Progress to 14.5 while retaining 41.6 Observation Control and
96.4 Coherence. Thus Coherence should be interpreted jointly with both control
execution and state advancement, rather than as recovery evidence by itself.

\noindent\textbf{PM-RoPE Architecture.}
To validate the structural design of PM-RoPE (Sec.~\ref{sec:projective-memory-rope}), we ablate the camera geometry injection pathways. We compare our complete architecture (\textbf{Full}) with two variants that remove the camera geometry injection in query/key (\textbf{VO only}) and value/output (\textbf{QK only}) separately, alongside our dual-branch, WorldPlay-style reimplementation~\citep{sun2025worldplay}. This row is not the official HY-WorldPlay model reported in Table~\ref{tab:stevo-results}.

As shown in Table~\ref{tab:ablation-pmrope}, the complete PM-RoPE architecture outperforms the partial variants in both low-level perceptual similarity (LPIPS) and high-level semantic consistency measured by Qwen3-VL~\citep{bai2025qwen3vl} scoring (Appendix~\ref{app:qwen-eval}). PM-RoPE also outperforms the \textbf{Dual Attention} baseline despite using a single attention operation, empirically supporting our analysis of decoupled branches under occlusion (Appendix~\ref{app:math-proof}). These high-level consistency gains align with the stronger State Progress and Physical Plausibility observed in the main STEVO-Bench results.
Appendix~\ref{app:efficiency} further measures this trade-off: PM-RoPE preserves
the vanilla KV-cache size, whereas the matched dual-attention reimplementation
uses a $2\times$ cache and higher streaming latency.

\begin{table}[h]

\centering
\small
\setlength{\tabcolsep}{4pt}
\caption{Ablation of 1.3B PM-RoPE structural variants. All rows use the same
initialization, data, curriculum, training budget, and evaluation set.
``Dual Attention'' is our WorldPlay-style \texttt{dual\_prope}
reimplementation, not the official HY-WorldPlay checkpoint. Metrics are
averaged across standard and noisy-memory retrieval conditions.}
\begin{tabular}{l c c c c c}
\toprule
\raisebox{0.75em}{Architecture Mode} & \raisebox{0.75em}{LPIPS $\downarrow$} & \shortstack{Degradation \\ Visible $\uparrow$} & \shortstack{Post-Degradation \\ Coherence $\uparrow$} & \shortstack{Temporal \\ Consistency $\uparrow$} & \shortstack{Content \\ Preservation $\uparrow$} \\
\midrule
Dual Attention (ours) & 0.507 & 4.0 & 2.9 & 2.7 & 3.3 \\
QK only & 0.508 & 3.5 & 3.2 & 2.5 & 2.9 \\
VO only & 0.452 & 4.0 & 3.8 & 3.5 & 3.7 \\
\textbf{Full (Ours)} & \textbf{0.403} & \textbf{4.1} & \textbf{3.9} & \textbf{3.8} & \textbf{3.9} \\
\bottomrule
\end{tabular}
\label{tab:ablation-pmrope}
\end{table}

\begin{table}[h]
\centering
\small
\setlength{\tabcolsep}{4pt}
\caption{Historical 1.3B dynamic-loss comparison. Regime probabilities match,
but source sampling weights differ; the result should not be interpreted as a
one-factor loss ablation.}
\label{tab:ablation-dynamic}
\begin{tabular}{l c c c c c}
\toprule
\raisebox{0.75em}{Loss Variant} & \raisebox{0.75em}{LPIPS $\downarrow$} & \shortstack{Degradation \\ Visible $\uparrow$} & \shortstack{Post-Degradation \\ Coherence $\uparrow$} & \shortstack{Temporal \\ Consistency $\uparrow$} & \shortstack{Content \\ Preservation $\uparrow$} \\
\midrule
Flow only & \textbf{0.395} & 2.6 & 3.1 & 3.3 & 3.1 \\
\textbf{Dynamic + adaptive (Ours)} & 0.424 & \textbf{3.4} & \textbf{3.5} & \textbf{3.6} & \textbf{3.6} \\
\bottomrule
\end{tabular}
\end{table}

\noindent\textbf{Dynamic Loss.}
We additionally report the historical dynamic-loss comparison used during
method development. The \textbf{Flow only} run disables adaptive temporal-delta
weighting, whereas \textbf{Dynamic + adaptive} enables it. Both retain the same
regime probabilities, but their stored configurations use different source
sampling weights; consequently, Table~\ref{tab:ablation-dynamic} is
descriptive and is not a clean one-factor causal ablation. The adaptive run
improves all Qwen3-VL~\citep{bai2025qwen3vl} semantic dimensions, while the
flow-only run obtains lower LPIPS, consistent with LPIPS favoring conservative
or frozen predictions. A matched loss-only rerun is required for strict
attribution.

\subsection{Attention Map Analysis for Dynamic Memory}
\label{sec:kv-importance}
\begin{figure}[t]
    \centering
    \includegraphics[width=0.24\linewidth]{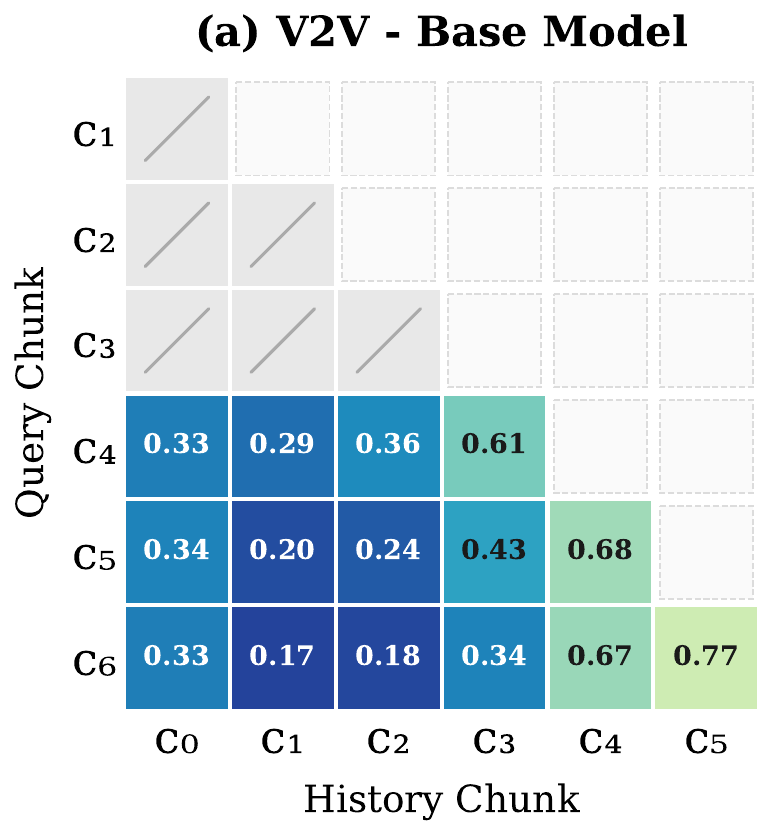}
    \includegraphics[width=0.24\linewidth]{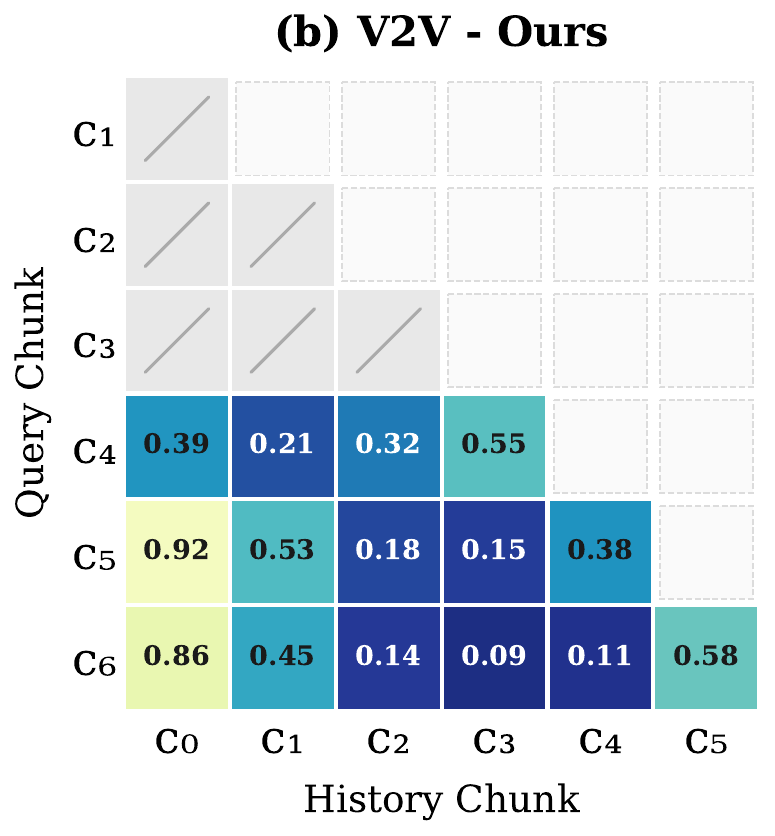}
    \includegraphics[width=0.24\linewidth]{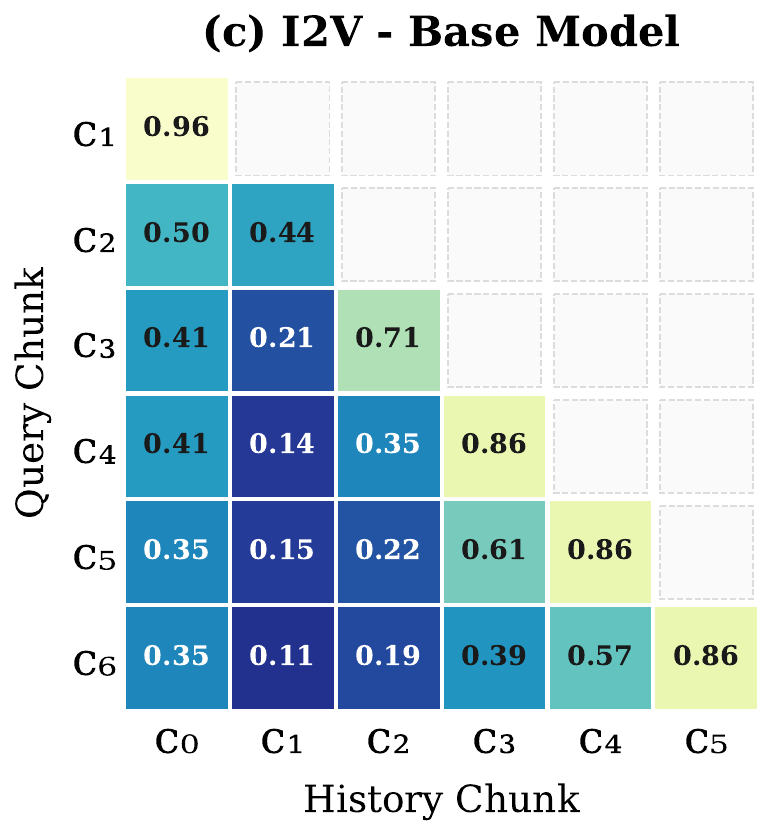}
    \includegraphics[width=0.24\linewidth]{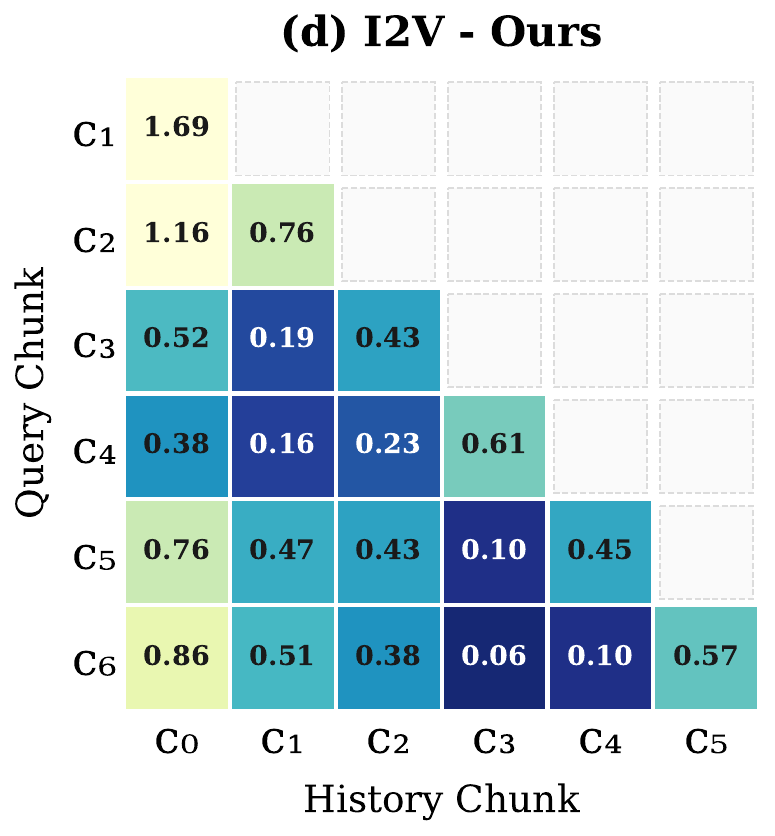}
    \caption{Selected KV-importance heatmaps (maximum over layers 20--24).
    Rows are query chunks and columns are history chunks; slashed cells in
    (a,\,b) are unavailable in V2V. These examples illustrate different
    attention patterns but do not establish population-level or causal memory
    use.}
    \label{fig:kv-importance}
\end{figure}

We use KV-importance maps to illustrate attention routing in selected V2V
occlusion and I2V camera-loop cases (Fig.~\ref{fig:kv-importance}). In these
examples, the base model concentrates attention near the diagonal. Under
occlusion, our model assigns high displayed importance to the pre-occlusion
anchor $c_0$ for queries $c_5$/$c_6$ (0.92/0.86); in the camera-loop example,
attention shifts toward $c_0$/$c_1$ when the camera returns. Because the figure
selects cases and takes a maximum over layers 20--24, it is a qualitative
diagnostic consistent with---but not proof of---historical-anchor retrieval.

\subsection{Qualitative Visualizations}
\label{sec:qualitative}

We present qualitative visualizations in Fig.~\ref{fig:comparison} covering
occlusion, camera motion, and re-observation after a positional gap. Together
with the KV-importance heatmaps (Sec.~\ref{sec:kv-importance}), these selected
cases are consistent with retrieval of historical evidence, but neither
attention visualization nor selected samples establish general long-gap state
evolution. Additional cases and a selected success gallery are provided in
Appendix~\ref{app:additional-qualitative}.

\section{Conclusion and Limitation}
\label{sec:limitations}
We present \ourmethod{}, empowering autoregressive video transformers to use their KV cache as dynamic memory for out-of-sight state evolution. By training on frame graphs with protected anchors, degraded intervals, and temporal gaps, \ourmethod{} learns to recover from reliable historical states instead of merely recent context. PM-RoPE further provides camera-aware spatiotemporal addressing while preserving pretrained attention pathways through one attention path. Experiments support short-horizon recovery under occlusion and darkness; camera-lookaway and long-duration state evolution remain challenging.

Matched controls refine this attribution: memory-oriented data teaches much of
the core duration-conditioned recovery behavior, while the complete
curriculum/PM-RoPE stack is most beneficial when retrieval must remain stable
across viewpoint change. In a 24-case V2V intervention, removing the clean
pre-interruption anchor reduces recovery PSNR, showing a measurable anchor
contribution in that controlled protocol.

\noindent\textbf{Limitation.} The strongest evidence supports short-horizon
out-of-sight recovery under occlusion and darkness rather than a general
symbolic world state or unlimited-duration physical simulation. Camera
lookaway and loop closure are evaluated, but their small subsets and weaker
control/progress scores do not support the same claim strength. The
duration probe further shows that elapsed-time evidence comes from rendered
event content: an otherwise empty RoPE gap preserves an anchor but does not by
itself advance state. The accompanying long-gap state sweep is therefore a
negative result, not evidence of physical simulation over an unobserved
duration. Finally, the pipeline relies on camera and depth quality; pose errors
in web videos may weaken PM-RoPE supervision and bias training toward sequences
with reliable tracks.

\noindent\textbf{Broader Impact.} Like other video generators, \ourmethod{} inherits risks around misleading media and dataset bias. We mitigate these through documented provenance, safety filtering, usage guidelines, a non-commercial research license, and release of restricted media as identifiers, manifests, annotations, or derived metadata rather than repackaged raw videos. Positively, \ourmethod{} also offers a diagnostic lens for hidden-state evolution, helping identify when visually plausible models fail to maintain persistent world states.

\bibliographystyle{ACM-Reference-Format}
\bibliography{main}

\clearpage
\appendix
\section{Additional Qualitative Cases}
\label{app:additional-qualitative}

\begin{figure*}[ht]
\centering
\includegraphics[width=\textwidth]{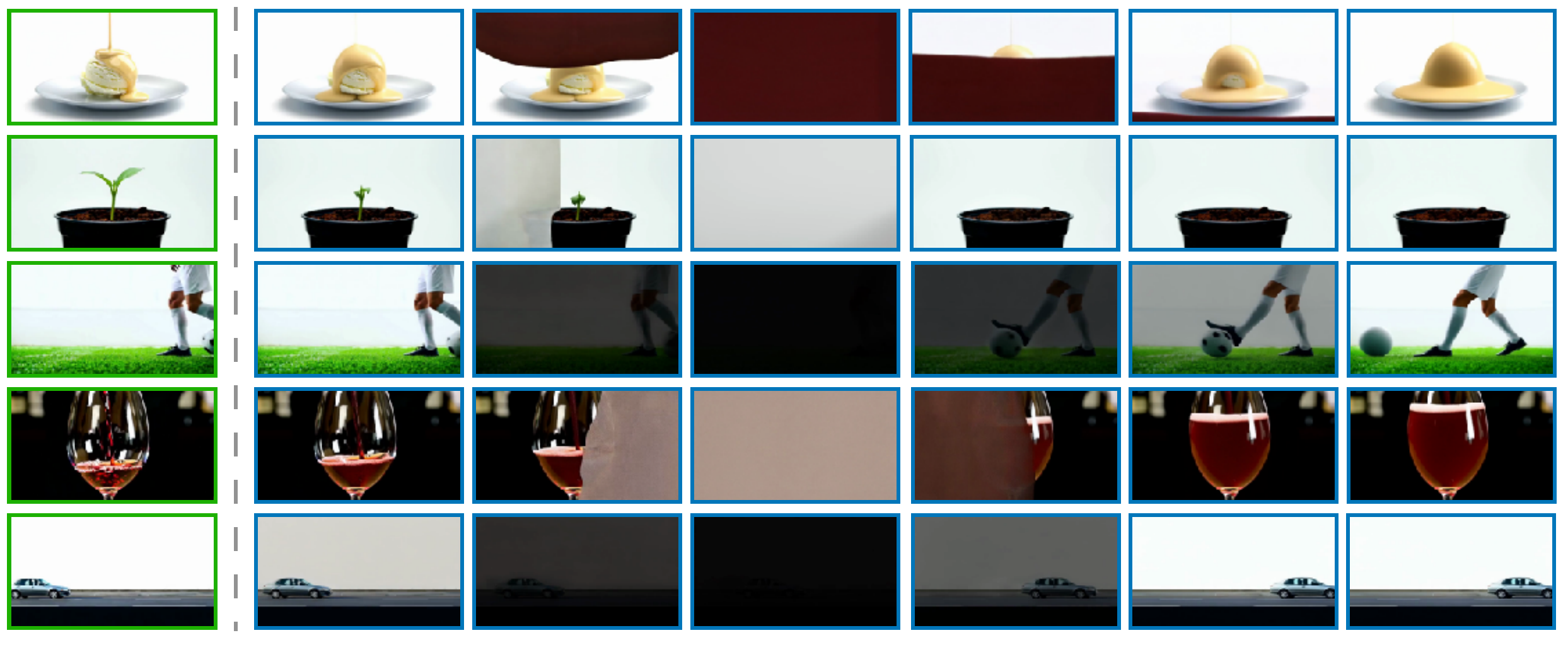}
\vspace{-2em}
\caption{Selected Image-to-Video (I2V) recovery examples spanning occlusion
and lighting-change prompts. These examples are illustrative; aggregate
behavior is reported in the quantitative evaluation.}
\label{fig:app-i2v}
\end{figure*}

\begin{figure*}[ht]
\centering
\includegraphics[width=\textwidth]{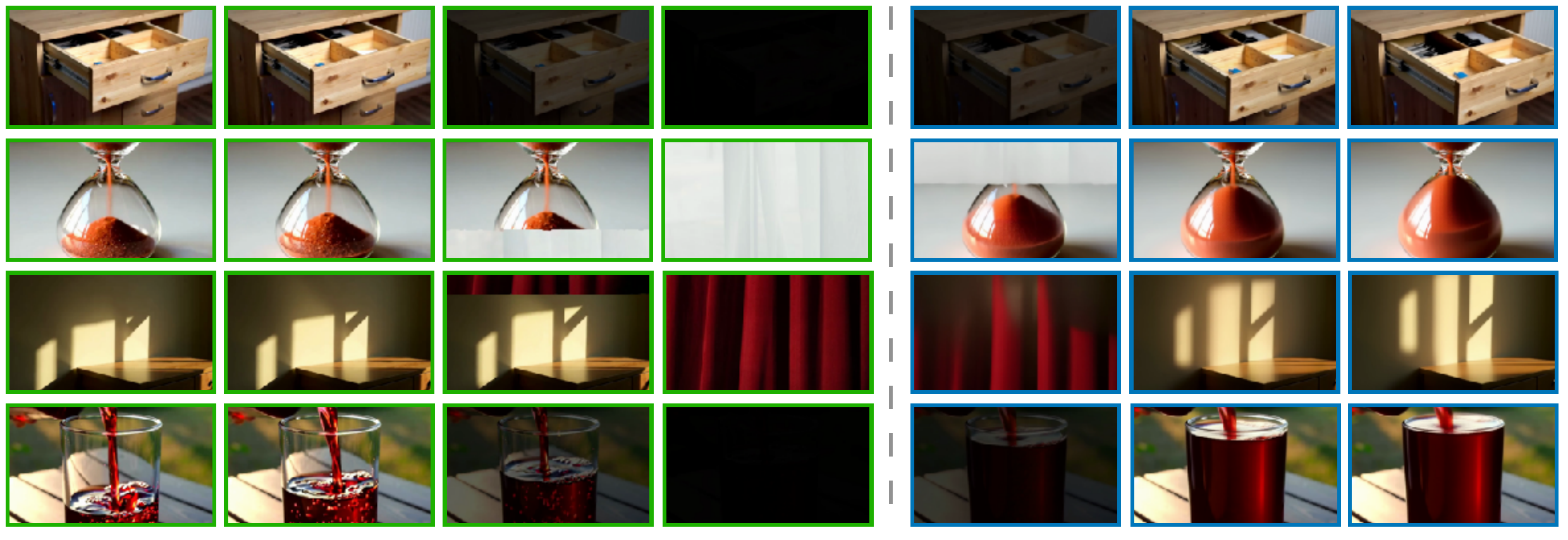}
\vspace{-2em}
\caption{Selected Video-to-Video (V2V) recovery examples conditioned on
provided prefixes.}
\label{fig:app-v2v}
\end{figure*}

\begin{figure*}[ht]
\centering
\includegraphics[width=\textwidth]{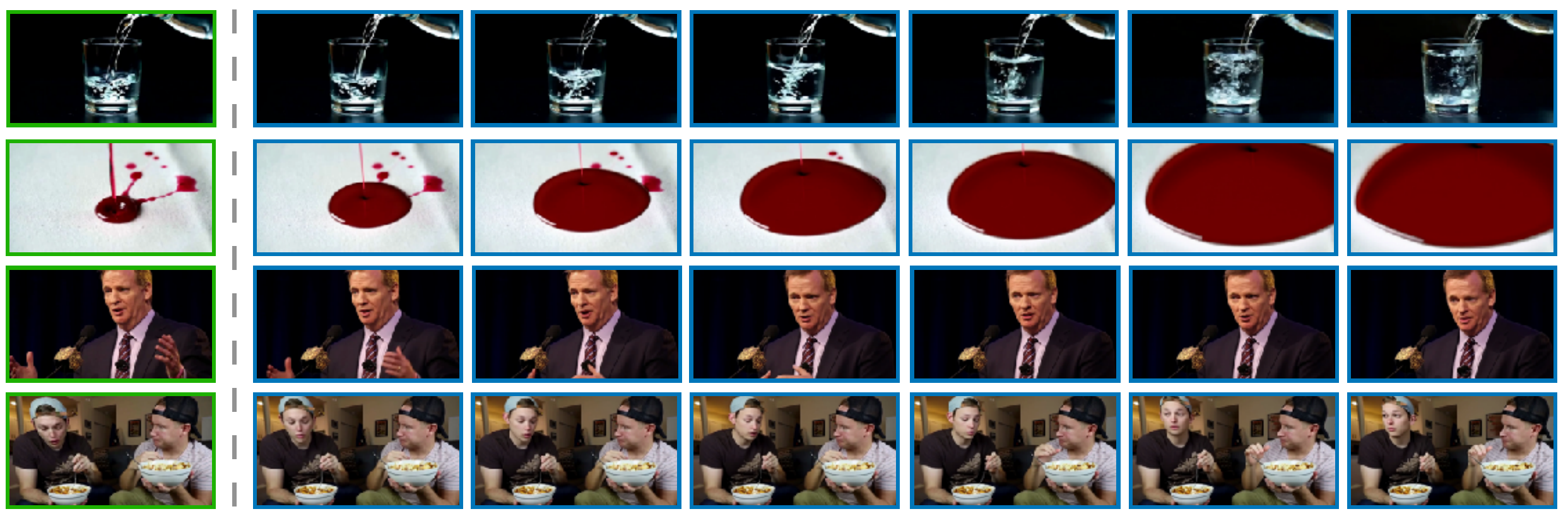}
\vspace{-2em}
\caption{Selected general Image-to-Video (I2V) generations from static input
images.}
\label{fig:app-i2v-general}
\end{figure*}

\begin{figure*}[t]
\centering
\includegraphics[width=\textwidth]{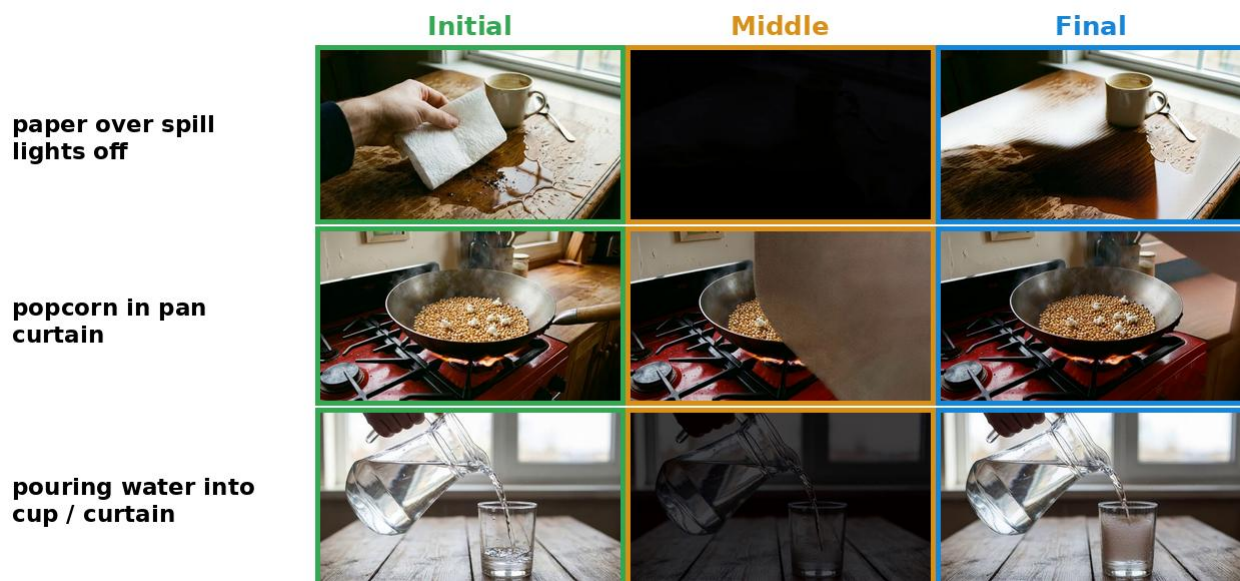}
\caption{Selected joint-success cases from our 221-video official STEVO-Bench
submission. Each displayed case receives positive official binary judgments
for both Observation Control and State Progress and is selected for visual
legibility. We show frames 0, 40, and 80; green, orange, and blue borders
indicate the initial, middle, and final frames, respectively. Frame 40 is not
asserted to be the event peak for every task. These selected successes are not
representative of the aggregate submission; Table~\ref{tab:stevo-results}
remains the quantitative summary.}
\label{fig:stevo-selected-gallery}
\end{figure*}

We provide additional qualitative examples across various settings. These
include Image-to-Video (I2V) recovery cases under occlusions and lighting
changes (Fig.~\ref{fig:app-i2v}), Video-to-Video (V2V) generation
(Fig.~\ref{fig:app-v2v}), and general I2V content generation
(Fig.~\ref{fig:app-i2v-general}). These coverage-oriented figures are selected
to span distinct settings; quantitative claims are computed on the complete
evaluation manifests. Figure~\ref{fig:stevo-selected-gallery} additionally
shows selected joint-success cases. In the
coverage-oriented figures, \textcolor{green}{green boxes} denote given
conditioning frames and \textcolor{blue}{blue boxes} denote generated frames.
This figure is illustrative rather than aggregate evidence. The archival
artifact package will include the complete 221-video submission, output map,
and per-case evaluation records needed to audit the sample.

\section{Dynamic Auxiliary Loss Details}
\label{app:dynamic-loss}

A core challenge in training out-of-sight state evolution is the imbalance of motion gradients. In localized dynamic events, such as water pouring into a cup within a largely static room, the state change only accounts for a small fraction of the canvas. Standard flow-matching loss averages errors across all pixels, causing the gradient direction to be dominated by static background reconstruction rather than the critical evolving state. To rectify this, we introduce a \emph{Dynamic Temporal-Delta Loss} $\mathcal{L}_{\Delta}$ paired with an \emph{Adaptive Weighting} strategy.

Alongside the main flow loss $\mathcal{L}_{\text{flow}}$ predicting the noise vector, we enforce an auxiliary penalty on the reconstructed clean latent $\hat{x}_0$ to match the ground-truth inter-frame changes $\Delta x_0$:
\begin{equation}
    \mathcal{L}_{\Delta} = \mathbb{E}_{t, \epsilon} \left[ \| (\hat{x}_{0,i} - \hat{x}_{0,i-1}) - (x_{0,i} - x_{0,i-1}) \|^2 \right],
\end{equation}
where $i$ denotes the frame index within a chunk. The total objective is $\mathcal{L} = \mathcal{L}_{\text{flow}} + \lambda_{\text{adapt}} \mathcal{L}_{\Delta}$. Crucially, the weight $\lambda_{\text{adapt}}$ is dynamically adjusted based on the global temporal variance $\sigma_{\text{batch}}$ of the current batch:
\begin{equation}
    \lambda_{\text{adapt}} = \alpha \cdot \exp(-\gamma \cdot \sigma_{\text{batch}}),
\end{equation}
where $\sigma_{\text{batch}} = \operatorname{Var}(\Delta x_0)$ computed over all spatial and temporal dimensions in the batch. We set $\alpha=0.2$ and $\gamma=5.0$. When global spatial change is small (e.g., static background with localized evolution), $\lambda_{\text{adapt}}$ boosts the delta loss to provide a strong optimization signal for the hidden state. Conversely, for sequences with massive overall spatial variation (e.g., fast camera pans), the weight automatically decays, returning optimization priority to the standard flow loss to preserve structural stability. This auxiliary loss is enabled after the first 2,000 iterations of finetuning to ensure the model first stabilizes on the primary flow objective.

\section{Identifiability of Dynamic Memory under Same-Position Occlusion}
\label{app:math-proof}

This section clarifies the mathematical limitation we target.  We do not claim
that RoPE~\citep{su2024roformer} forbids long-range retrieval: for a query at time $t_i$ and a key at
time $t_j$, standard RoPE gives:
\begin{equation}
  \langle R(t_i)q_i, R(t_j)k_j\rangle
  = q_i^\top R(t_j-t_i)k_j,
\end{equation}
so the score depends on the relative temporal offset, but it is not generally a
monotone function of distance.  The issue is instead \emph{identifiability}: a
decoupled spatial-memory branch and a separately reframed temporal branch
(like WorldPlay~\citep{sun2025worldplay}) do not
provide a single key for ``the last clean state at this spatial address before a
degradation boundary.''

\subsection{Problem Setup}
Consider a recovery query $q$ after an occlusion is removed.  Let $p$ denote the
spatial address or camera ray of the dynamic region in the recovered view.  There
are two relevant cached candidates:
\begin{itemize}
    \item \textbf{Clean anchor ($a$)}: the last visible state before degradation,
    at time $t_a$ and spatial address $p$.
    \item \textbf{Corrupted context ($c$)}: a more recent frame during degradation,
    at time $t_c$ and the same spatial address $p$, but whose visible content is
    the occluder or darkness rather than the underlying state.
\end{itemize}
The times satisfy $t_a < t_c < t_q$.  The desired recovery edge is not simply
``same spatial address'' or ``nearest previous time'', but the conjunction
\begin{equation}
  j^\star
  =
  \arg\max_{j<t_q}
  \mathbf{1}\{s_j=s_q\}
  \mathbf{1}\{t_j < \tau_{\mathrm{deg}}\}
  \rho(x_j,x_q),
  \label{eq:clean-anchor-selector}
\end{equation}
where $s_j$ is the spatial/camera address, $\tau_{\mathrm{deg}}$ is the start of
the degradation interval, and $\rho$ measures visual or state compatibility.  In
words, the key must jointly encode spatial alignment and whether the candidate is
the clean pre-degradation state.

\subsection{Decoupled Spatial and Temporal Scores}
Many memory-augmented generators use separate mechanisms for spatial and
temporal routing.  Abstractly, a decoupled score has the form:
\begin{equation}
  \ell(q,j)
  =
  \phi_{\mathrm{sp}}(s_q,s_j)
  +
  \phi_{\mathrm{tmp}}(\pi_q,\pi_j)
  +
  \phi_{\mathrm{cnt}}(x_q,x_j),
  \label{eq:decoupled-score}
\end{equation}
where $\phi_{\mathrm{sp}}$ is a spatial or camera-address score, $\phi_{\mathrm{tmp}}$
is a temporal-position score, and $\phi_{\mathrm{cnt}}$ is a content score.  In
WorldPlay-style reconstituted context memory~\citep{sun2025worldplay}, the
selected historical frames are written into compact cache slots.  Therefore the
temporal position $\pi_j$ used by the RoPE branch is a \emph{cache order} rather
than the original absolute time or an explicit degradation-boundary coordinate.

For the clean anchor $a$ and corrupted context $c$, same-position occlusion gives:
\begin{equation}
  s_a=s_c=s_q
  \quad \Rightarrow \quad
  \phi_{\mathrm{sp}}(s_q,s_a)=\phi_{\mathrm{sp}}(s_q,s_c).
  \label{eq:spatial-tie}
\end{equation}
Thus a purely spatial or PRoPE-style branch cannot distinguish whether the
same-ray cached entry is a clean state or an occluder.  The decision is reduced
to temporal cache order and content:
\begin{equation}
  \ell(q,a)-\ell(q,c)
  =
  \bigl[\phi_{\mathrm{tmp}}(\pi_q,\pi_a)-\phi_{\mathrm{tmp}}(\pi_q,\pi_c)\bigr]
  +
  \bigl[\phi_{\mathrm{cnt}}(x_q,x_a)-\phi_{\mathrm{cnt}}(x_q,x_c)\bigr].
  \label{eq:decoupled-difference}
\end{equation}
Neither term explicitly contains the predicate $t_j<\tau_{\mathrm{deg}}$ from
Eq.~\ref{eq:clean-anchor-selector}.  If the content term is ambiguous or
uninformative--for example, the query region is currently unobserved, blacked
out, or covered by an occluder--then the score has no identifiable variable that
selects the clean pre-degradation anchor over the more recent corrupted context.

\subsection{Implication}
The limitation is therefore not a theorem that RoPE must always prefer nearby
tokens.  Rather, a decoupled spatial-temporal memory lacks the joint address
needed for hidden-state recovery: spatial routing can find the same ray, but not
the clean state boundary; compact temporal routing can preserve local order, but
not the original elapsed-time relation to the degradation event.  \ourmethod{}
addresses this by writing memory nodes with their original RoPE positions and
camera metadata, and by training reference-cache and node-drop regimes in which
the correct recovery depends on retrieving the clean anchor rather than the
recent degraded context.

\section{VLM Evaluation Details}
\label{app:qwen-eval}

To comprehensively evaluate the state-recovery capabilities in our ablation studies, we utilize Qwen3-VL~\citep{bai2025qwen3vl} to provide high-level semantic scoring. For each generated suffix, we prompt the VLM to act as an expert video reviewer and score the video from 1 (worst) to 5 (best) across four dimensions: Overall Quality, Temporal Consistency, Content Preservation, and Artifact Suppression (which correlates to the inverse of Degradation Visible).

\textbf{Evaluation Protocol.} We evaluate 200 clips randomly sampled from the recovery evaluation set across all degradation conditions. To ensure a blind and unbiased assessment, videos from different model variants were shuffled and presented to the LMM without identifying tags. Each video was evaluated three times using a temperature of 0.7 to account for stochasticity in the LMM's reasoning, with the final scores averaged across these trials. We used a fixed random seed of 42 for all sampling and inference steps. Tables~\ref{tab:ablation-pmrope}--\ref{tab:ablation-dynamic} report point estimates and do not display confidence intervals; the archival artifact package will include per-example scores and the bootstrap script.

The standard prompt template provided to the model follows the structure below, which asks the VLM to observe the transition from the visible history context to the newly recovered state:

\begin{quote}
\textit{``You are an expert video reviewer evaluating the physical consistency of generated videos after a period of occlusion or degradation. Observe the initial state of the objects and the background in the first few frames, and then closely examine the frames after the degradation ends.\\
Please rate the following dimensions on a scale of 1 to 5:\\
1. Overall Quality: Is the recovered video visually coherent and realistic?\\
2. Temporal Consistency: Do the objects resume their motion and trajectory naturally without abrupt jumps?\\
3. Content Preservation: Are the identities, shapes, and textures of the original subjects completely preserved upon recovery?\\
4. Artifact Suppression: Is the video free from flickering, structural morphing, or lingering textures from the occluder?''}
\end{quote}

These individual scores are averaged across the evaluation set to yield the quantitative results reported in the ablation tables.

\section{Protocol Definitions and Additional Quantitative Results}
\label{app:additional-results}

\subsection{I2V and V2V Protocols}

We distinguish two evaluation protocols throughout the revision. In
\emph{image-to-video} (I2V), the input is a first frame and an instruction; the
model must generate both the interruption and the subsequent recovery.
STEVO-Bench and VBench use this protocol. In \emph{video-to-video} (V2V), a
clean or synthetically degraded prefix through the interruption peak is given,
and the model generates the recovery suffix. Because the degradation is applied
to a complete source video, V2V provides a time-aligned clean suffix and thus
supports direct PSNR, SSIM, and motion-region measurements. Unless stated
otherwise, the controlled V2V experiments below use 24 fixed-seed cases and a
single generation seed per case. They are single-seed point estimates; no
generation-seed variance or confidence intervals are available for these
24-case probes. In both protocols, the submitted video is
decoded directly from the generator output: the inference path contains no
post-hoc compositor, frame replacement, or benchmark-specific
observation-control overlay.

\subsection{STEVO-Bench Composition, Metrics, and Aggregation}
\label{app:stevo-protocol}

The evaluation manifest contains 85 underlying projects expanded into 221
task variants: 69 use a cardboard occluder, 69 use a curtain occluder, 67 use
a lights-off interval, and 16 use camera lookaway/return control. The cases span
seven benchmark annotation levels. The benchmark paper describes 225 tasks;
our official submission/evaluator snapshot contains 221 scored variants. All
statistics in this paper refer to that exact 221-case snapshot, whose task
identifiers and manifest version will accompany the archival artifacts. Each
I2V case supplies an initial image and
an instruction that specifies both the underlying action or physical state
change and the visibility/control intervention; the future video is not
provided to the generator. Our submission videos contain 81 frames at 16 fps
and $480\times832$ resolution. Figure~\ref{fig:teaser}(a--c) shows
representative light-off, full-frame-occlusion, and camera-lookaway cases with
their initial, interrupted, and recovered frames.

The official evaluation reports five percentages. \emph{State Progress} asks
whether the underlying process reaches a measurably later state despite the
unobserved interval. \emph{Physical Plausibility} checks for physically
credible, artifact-free dynamics. \emph{Coherence} checks whether objects and
the scene remain temporally continuous rather than resetting or cutting.
\emph{Observation Control} checks whether the requested occlusion, lighting, or
camera intervention occurs, and \emph{Action Control} checks whether the
requested action is executed. If $z_{im}\in\{0,1\}$ is the official
case-level judgment for metric $m$ on case $i$, the displayed metric is
\begin{equation}
S_m=\frac{100}{221}\sum_{i=1}^{221}z_{im}.
\end{equation}
The Total column in Table~\ref{tab:stevo-results} is the unweighted sum
$\sum_m S_m$ and therefore ranges from 0 to 500. It is not an additional
judge. All baseline entries in that table are official leaderboard aggregates;
our entry is the official aggregate for the 1.3B full-data
\texttt{checkpoint-1000} submission. We generate it with 25 denoising steps
and seed 1000; each binary judgment is the majority of three
Gemini-2.5-Flash evaluations.

\subsection{Per-Dimension VBench Results and Dynamic-Degree Attribution}
\label{app:vbench-detailed}

Table~\ref{tab:vbench-detailed} expands the three VBench groups in
Table~\ref{tab:nocamera-results} into all eight constituent dimensions. The
three reported groups are equal-weight means: Semantic Accuracy uses I2V subject
and background; Motion \& Temporal uses subject consistency, background
consistency, motion smoothness, and dynamic degree; Visual Quality uses
aesthetic and imaging quality.

\begin{table}[H]
\centering
\small
\setlength{\tabcolsep}{4pt}
\caption{Per-dimension VBench I2V results under a common runner, seed set, and
resolution. The 5B data-only and full-method arms share initialization,
training data, 6,000 optimization steps, and 64 A100 GPUs. Data-only uses
standard RoPE and omits the memory curriculum.}
\label{tab:vbench-detailed}
\begin{tabular}{lccc}
\toprule
Dimension $\uparrow$ & InSpatio-World & Data-only & \ourmethod{} \\
\midrule
I2V subject & 89.25 & 97.70 & \textbf{97.72} \\
I2V background & 91.32 & 98.54 & \textbf{98.60} \\
Subject consistency & 76.59 & 81.26 & \textbf{81.62} \\
Background consistency & 87.34 & 89.23 & \textbf{89.44} \\
Motion smoothness & 98.62 & \textbf{99.16} & 99.15 \\
Dynamic degree & \textbf{41.87} & 38.61 & 38.67 \\
Aesthetic quality & 54.90 & 60.12 & \textbf{60.36} \\
Imaging quality & 58.76 & \textbf{70.09} & 70.00 \\
\bottomrule
\end{tabular}
\end{table}

The 1.3B baseline and 5B full model differ on seven of eight dimensions, but
that cross-scale comparison is descriptive rather than causal. Within the
matched 5B trained arms, data-only and full differ by at most 0.4 points on
every dimension, and both have nearly identical Dynamic Degree (38.61 versus
38.67). Thus the two matched arms have similar point estimates; without paired
uncertainty or a prespecified margin, we do not claim equivalence. In the
cross-scale comparison, 96
samples change from dynamic to static and 57 from static to dynamic; these
counts should not be attributed to PM-RoPE or the curriculum because model size
and training data also differ.

\subsection{Matched-Compute Data and Generalization Controls}
\label{app:matched-controls}

We isolate the contribution of memory-oriented data using matched 5B control
runs. Data-only and full \ourmethod{} use the same pretrained initialization,
complete training mixture, 6,000 steps, 64 A100 GPUs, and $256\times448$
training/evaluation resolution. The data-only arm uses standard RoPE and no
memory curriculum. A leave-one-family-out arm removes all light-toggle examples
from training.

\begin{table}[H]
\centering
\small
\setlength{\tabcolsep}{4pt}
\caption{Matched-compute V2V recovery against the time-aligned clean suffix.
Fog is absent from both the training degradation families and STEVO-Bench;
the no-light arm tests transfer to a held-out training family. Each cell uses
24 fixed-seed cases.}
\label{tab:matched-v2v}
\begin{tabular}{lccc}
\toprule
Model & Fog PSNR/SSIM & Light PSNR/SSIM &
\shortstack{Light + camera pan\\PSNR/SSIM} \\
\midrule
Raw Wan2.2-5B & 8.5 / 0.31 & 7.6 / 0.24 & --- \\
Data-only & 19.3 / 0.92 & 17.6 / 0.90 & 16.6 / 0.82 \\
No-light training & 19.7 / 0.92 & \textbf{17.8 / 0.91} & --- \\
\textbf{\ourmethod{}} & \textbf{19.8 / 0.92} &
\textbf{17.8 / 0.91} & \textbf{18.3 / 0.87} \\
\bottomrule
\end{tabular}
\end{table}

The static light result shows that the data teaches much of the core recovery
behavior. Under viewpoint change, however, the complete method improves over
data-only by 1.7 dB PSNR and 0.05 SSIM, consistent with the combined
curriculum/PM-RoPE stack improving retrieval across camera motion. Removing the
light family from training does not reduce light recovery at the reported
precision, while the held-out fog condition remains far above the raw model.
These controls support degradation-family transfer, while not by themselves
separating curriculum from PM-RoPE.

\begin{table}[H]
\centering
\small
\caption{Production-checkpoint V2V fog-recovery slice. Both rows use the 5B
backbone, 25 denoising steps, the same 24 held-out prefixes, a fixed seed, and
a time-aligned clean suffix. Motion-region PSNR is computed inside the
ground-truth motion mask.}
\label{tab:production-v2v}
\begin{tabular}{lccc}
\toprule
Checkpoint & Full PSNR $\uparrow$ & Full SSIM $\uparrow$ &
Motion PSNR $\uparrow$ \\
\midrule
Raw Wan2.2-5B initialization & 10.25 & 0.484 & 9.789 \\
\ourmethod{} teacher & \textbf{19.93} & \textbf{0.923} & \textbf{15.592} \\
\bottomrule
\end{tabular}
\end{table}

Table~\ref{tab:production-v2v} is distinct from the matched-compute controls in
Table~\ref{tab:matched-v2v}: it uses production checkpoints and therefore
should not be used for data-versus-method attribution.

\subsection{Duration-Conditioned State Advance}
\label{app:duration-probe}

We fix the visible prefix and vary only the length $K$ of a rendered dark
interval from one to four chunks. Recovery is measured at a fixed offset after
the lights return. The objective statistic
\[
\Delta(K)=\operatorname{SSIM}(x^{\mathrm{rec}}_K,x^{\mathrm{GT}}_K)
-\operatorname{SSIM}(x^{\mathrm{rec}}_K,x^{\mathrm{GT}}_{\mathrm{pre}})
\]
is positive when recovery is closer to the time-matched state than to the
pre-interruption state.

\begin{table}[H]
\centering
\small
\caption{Objective duration-conditioned state advance on 24 V2V cases using
the 5B backbone and one fixed generation seed. The percentage is the fraction
with $\Delta(4)>0$ across cases, not uncertainty over seeds.}
\label{tab:duration-objective}
\begin{tabular}{lccccc}
\toprule
Model & $K=1$ & $K=2$ & $K=3$ & $K=4$ & Positive at $K=4$ \\
\midrule
Raw Wan2.2-5B & $-0.001$ & $+0.007$ & $+0.015$ & $+0.014$ & 62.5\% \\
\ourmethod{} teacher & $+0.068$ & $+0.112$ & $+0.141$ & $+0.144$ & 95.8\% \\
\ourmethod{} 4-step & $+0.079$ & $+0.118$ & $+0.150$ & $\mathbf{+0.164}$ & 95.8\% \\
\bottomrule
\end{tabular}
\end{table}

The objective curve increases with hidden duration for both the teacher and
four-step student. Independent Gemini and Qwen judges agree in direction,
although the objective metric resolves the short-duration differences more
reliably. A separate RoPE-offset-only probe retains the anchor but produces
approximately flat state advance. Thus, the evidence supports a narrower
mechanism: rendered event content supplies elapsed-time evidence, while
PM-RoPE and the cache provide robust anchor retrieval. The current checkpoint
does not convert an otherwise empty positional gap into extrapolated physical
time.

\subsection{Causal Anchor Knockout and Long-Gap Retrieval}
\label{app:anchor-probes}

To quantify the contribution of the pre-interruption image, we remove only the
clean anchor from the cache while keeping the black history chunks, latent
inputs, positions, prompt, and seed fixed. In this 24-case, one-seed V2V
protocol, Table~\ref{tab:anchor-knockout} shows that anchor removal reduces
recovery PSNR by 2.0 dB. The intervention establishes a contribution within
this protocol; it does not by itself rule out other conditioning pathways.

\begin{table}[H]
\centering
\small
\caption{Anchor-cache intervention on 24 black-interruption V2V cases using
the 5B four-step student.}
\label{tab:anchor-knockout}
\begin{tabular}{lcc}
\toprule
Cache contents & PSNR $\uparrow$ & SSIM $\uparrow$ \\
\midrule
Anchor + dark history & \textbf{18.32} & \textbf{0.909} \\
Anchor only & 17.04 & 0.863 \\
Dark history only (anchor removed) & 16.34 & 0.881 \\
\bottomrule
\end{tabular}
\end{table}

We additionally place the recovery query at increasingly distant,
non-contiguous RoPE positions while keeping the stored cache size fixed.

\begin{table}[H]
\centering
\small
\setlength{\tabcolsep}{3.5pt}
\caption{Retrieval and state behavior as the positional gap $G$ increases for
the 5B four-step student on 24 cases and one fixed generation seed. Anchor mass is mean recovery attention
assigned to the original clean anchor. State advance is the Qwen3-VL
process-progress score of the recovery frame minus that of the visible prefix.
The 12--25 gap range was configured for reference-cache training but inactive
in this checkpoint, making this a retrieval diagnostic rather than evidence of
trained gap extrapolation.}
\label{tab:gap-attention}
\begin{tabular}{lrrrrrrrrrr}
\toprule
$G$ & 1 & 2 & 4 & 8 & 12 & 16 & 25 & 32 & 40 & 60 \\
\midrule
Anchor mass & .834 & .818 & .812 & .858 & .849 & .839 & .745 & .879 & .834 & .694 \\
State advance & +3.1 & +4.4 & +3.3 & $-2.3$ & $-0.4$ & $-2.3$ & +2.1 & $-1.3$ & $-6.7$ & $-1.0$ \\
\bottomrule
\end{tabular}
\end{table}

At $G=60$, \ourmethod{} stores 5,850 tokens, compared with 73,710 tokens for a
naive stream retaining every intervening chunk (12.6$\times$ more). Across four
repeated interruptions, attention mass on the original anchor is
0.849/0.486/0.387/0.401: newer recoveries absorb some attention, but the
original scene witness remains addressable. This probe measures attention
routing and cache cost. In contrast, per-case progress versus $G$ has mean
Spearman correlation $-0.19$ (median $-0.17$), and the aggregate state-advance
row remains approximately flat around zero. Thus long-range addressability
survives, but an empty positional gap neither supplies elapsed-time evidence nor
improves recovery-state advancement.

\subsection{Measured PM-RoPE Efficiency}
\label{app:efficiency}

Table~\ref{tab:efficiency} reports a controlled single-A100 inference benchmark
at $480\times832$, 81 frames, and bfloat16. Each rollout contains seven chunks,
with four denoising calls and one context-write call per chunk; values are
medians over five post-warmup runs. Dual attention is our matched
WorldPlay-style reimplementation used in Table~\ref{tab:ablation-pmrope}, not
the released HY-WorldPlay model.

\begin{table}[H]
\centering
\small
\setlength{\tabcolsep}{3pt}
\caption{Measured streaming efficiency. KV cache is measured at full 21-latent-
frame history; FLOPs are analytic self-attention FLOPs per block at the final
chunk.}
\label{tab:efficiency}
\begin{tabular}{llrrrr}
\toprule
Backbone & Mode & Time (s) & Peak GB & KV GB & GFLOPs/block \\
\midrule
1.3B & Vanilla RoPE & 7.48 & 9.25 & 6.04 & 1030 \\
1.3B & PM-RoPE & 9.00 & 9.69 & \textbf{6.04} & 1075 \\
1.3B & Dual attention & 12.67 & 15.77 & 12.08 & 1973 \\
\midrule
5B & Vanilla RoPE & 3.95 & 13.22 & 3.02 & 206 \\
5B & PM-RoPE & 4.88 & 14.42 & \textbf{3.02} & 250 \\
5B & Dual attention & 6.05 & 16.97 & 6.04 & 324 \\
\bottomrule
\end{tabular}
\end{table}

PM-RoPE leaves KV-cache size unchanged relative to vanilla RoPE, whereas the
dual branch duplicates it. Dual attention is 1.41$\times$/1.24$\times$ slower
than PM-RoPE on the 1.3B/5B backbones, respectively. Its latency also grows
faster with retained history (54 versus 30 ms per added chunk at 1.3B), because
the second attention pass rereads the growing duplicated cache. These are
inference-only measurements on one A100; they do not estimate distributed
training cost. PM-RoPE itself is 1.20$\times$/1.24$\times$ slower than vanilla
RoPE in the 1.3B/5B implementations, so ``one attention path'' does not mean
zero overhead. Absolute values should only be compared within a backbone
because the two implementations use different token layouts and cache
dimensions.

\begin{table}[H]
\centering
\small
\caption{Main training data components.}
\begin{tabular}{p{0.24\linewidth} p{0.30\linewidth} p{0.36\linewidth}}
\toprule
Component & Source & Role \\
\midrule
General videos & OpenVid, DL3DV-10K & preserve appearance and camera motion \\
Natural dynamics & Pexels + VLM filtering & visible state evolution and interactions \\
Hard dynamics & generative video models & high-occupancy state changes \\
Camera loops & Pexels + Neoverse rendering & loop-closure memory supervision \\
\bottomrule
\end{tabular}
\label{tab:dataset-mix}
\end{table}

\begin{table}[H]
\centering
\small
\setlength{\tabcolsep}{5pt}
\caption{Concurrent source-sampling distribution for the 1.3B full-data run.
Weights sum to one; they are sampling probabilities, not sequential stages.}
\label{tab:source-weights}
\begin{tabular}{lr@{\qquad}lr}
\toprule
Source & Weight & Source & Weight \\
\midrule
OpenVid DA3 & 0.10 & DL3DV DA3 & 0.20 \\
Pexels DA3 & 0.16 & Helios pouring & 0.04 \\
Helios curriculum & 0.10 & Helios broad dynamics & 0.10 \\
Helios loop/pan & 0.05 & Helios STEVO-gap & 0.12 \\
Veo3 occlusion & 0.03 & Pexels pan/evolution & 0.10 \\
\bottomrule
\end{tabular}
\end{table}

\paragraph{Scale and exposure accounting.}
The weights in Table~\ref{tab:source-weights} are configured draw
probabilities, not realized exposure counts. The current artifact package does
not contain a frozen step-level exposure log for the historical 1.3B run, so
we cannot audit exact accepted unique clips/hours, per-source draw counts, or
cross-source deduplication for that run. Candidate-pool and public-release row
counts must therefore not be interpreted as the realized training size.
The archival artifact ledger must include the pinned input manifests, their
hashes and accepted-row counts, the deduplication and split unit, and a
realized per-source draw census before making a stronger scale claim.

\section{Ablation of Data Construction}
\label{app:ablation-data}

We study memory-oriented data construction through three \emph{separate}
trainings from the same initialization and for the same number of steps. Within
each run, all listed sources are sampled concurrently by a weighted sampler;
the rows are cumulative data mixtures, not sequential training stages. The
\emph{Base} variant uses general videos (OpenVid~\citep{nan2024openvid} and
DL3DV~\citep{ling2023dl3dv}). \emph{Real Dynamics \& Loops} adds Pexels and
NeoVerse-rendered camera loops, and \textbf{Full} additionally adds generated
hard dynamics.

Table~\ref{tab:ablation-data} is a descriptive comparison of cumulative
mixtures. The real-dynamics-and-loops row has the lowest LPIPS, while the full
mixture has the highest reported post-degradation coherence, temporal
consistency, and content-preservation scores but worse LPIPS. Because sources
are bundled and uncertainty is unavailable, the table does not isolate
NeoVerse or establish statistical significance for an individual source.

\begin{table}[H]
\centering
\small
\setlength{\tabcolsep}{4pt}
\caption{Cumulative 1.3B data-mixture ablation. Each row is a separate,
matched-budget run whose listed sources are mixed concurrently; rows are not
stages of one curriculum.}
\begin{tabular}{l c c c c c}
\toprule
\raisebox{0.75em}{Data Mixture} & \raisebox{0.75em}{LPIPS $\downarrow$} & \shortstack{Degradation \\ Visible $\uparrow$} & \shortstack{Post-Degradation \\ Coherence $\uparrow$} & \shortstack{Temporal \\ Consistency $\uparrow$} & \shortstack{Content \\ Preservation $\uparrow$} \\
\midrule
Base (General Videos) & 0.422 & 2.0 & 1.9 & 1.8 & 1.9 \\
+ Real Dynamics \& Loops & \textbf{0.363} & \textbf{2.2} & 2.2 & 2.0 & 2.3 \\
\textbf{+ Synthetic Hard Dynamics} & 0.515 & \textbf{2.2} & \textbf{2.4} & \textbf{2.2} & \textbf{2.6} \\
\bottomrule
\end{tabular}
\label{tab:ablation-data}
\end{table}

\section{Data Construction Details}
\label{app:data}
The main components of our training data are listed in Table~\ref{tab:dataset-mix}.

\subsection{Dynamic Taxonomy and Retrieval Policy}

We build a broad dynamic taxonomy covering over one hundred scene-event types.
Each type receives an importance score that reflects three factors: whether the
state change is visually salient, whether the event is common enough to collect
at scale, and whether it is useful for hidden-state recovery after interruption.
The high-priority groups include monotonic state changes such as
pouring/filling, burning, melting, growing, and diffusion, as well as
large-occupancy motion, deformation, lighting/weather changes, and
human-object interactions.

The taxonomy serves two roles.  For Pexels retrieval, it provides keyword
queries and negative filters.  For generated or VLM-annotated clips, it acts as
caption guidance: captions must describe the object, the state variable, and
the temporal direction of change.  After retrieval, Qwen3-VL~\citep{bai2025qwen3vl}
is used for secondary filtering.  It estimates whether the clip contains a
scene cut, which dynamic event is present, how much image area the dynamic
region occupies, and how long the event remains visible.  We keep
high-occupancy, temporally sustained dynamic clips and discard clips dominated
by static scenery, montage cuts, or tiny localized motion.

\begin{table}[H]
\centering
\small
\caption{Dynamic retrieval taxonomy.  The full internal taxonomy covers over
one hundred event types; this table groups them by retrieval and captioning
role.  Counts are from the tagged curation table and are not mutually
exclusive.}
\begin{tabular}{p{0.24\linewidth} r p{0.50\linewidth}}
\toprule
Tag group & Candidates & Retrieval keywords and caption cues \\
\midrule
Translational & 39{,}205 & walk/run, vehicle motion, fall/drop, fly/float; caption moving object, direction, and speed \\
Monotonic state & 20{,}076 & pouring/filling, burn, melt, grow/bloom, diffuse/spread; caption state variable and monotonic change \\
Deformation & 13{,}793 & dance/yoga, body gesture, fabric motion, bubbles; caption shape change and actor/object \\
Lighting/weather & 11{,}745 & sun, rain/snow, wind/fog, cloud shadow, light switch; caption illumination or weather transition \\
Interaction & 10{,}218 & cooking, mixing, assembly, ball sports, tool use; caption actor-object interaction and outcome \\
Periodic & 3{,}210 & mechanical rotation, biological cycles, water periodic motion; caption periodic source and phase \\
Discrete & 1{,}147 & doors, lids, page flips, switches; caption trigger, before-state, and after-state \\
\bottomrule
\end{tabular}
\label{tab:dynamic-tags}
\end{table}

\subsection{Event Nodes and Degradations}

For each selected clip, we construct a clean video, an optional degraded video,
a caption, and a set of latent chunk nodes used by the memory curriculum.  The
main event families are camera loops, light toggles, moving occluders, and zoom
or camera perturbations.  Camera-loop nodes are derived from the trajectory:
the memory anchor is the initial view, and the recovery node is the later frame
whose camera pose returns nearest to that view.  Light-toggle nodes are derived
from the dimming schedule, with anchors before the dark interval and recovery
nodes after relighting.  Occluder nodes are derived from the enter, full-cover,
and exit frames.  Zoom nodes are derived from peak crop scale and recovery
frames, with intrinsics updated to match the crop.

\paragraph{Targeted return-pose sanity check.}
For the 12 fixed \texttt{pan\_yaw\_8dir} trajectories used in our project-page
and evaluation examples, we compute rotation relative to frame 0 and define a
return as re-entering $\max(2^\circ,0.1\,r_{\max})$ after the interior rotation
peak. All 12 trajectories peak at frames 38--42 with
$r_{\max}=14.05^\circ$--$35.56^\circ$, re-enter the return threshold at frames
63--65, and end with only $0.04^\circ$--$1.20^\circ$ residual rotation.
Nonzero camera-center residuals remain, so this verifies orientation return for
these 12 trajectories rather than exact pixel closure or corpus-wide pose
quality. We therefore define return supervision using pose overlap and protect
the initial witness instead of requiring the final camera center to equal the
initial center. This selected-example sanity check is not a substitute for a corpus-wide
retention statistic or a real-versus-rendered loop ablation.

All sources are normalized into the same training format.  Clips are sampled as
81 frames at 16 fps and resized to the training resolution ($480\times832$ in
our main runs).  After VAE temporal compression, this gives 21 latent frames,
which we group into seven chunks of three latent frames.  Intrinsics are scaled
to the training resolution; extrinsics are stored as OpenCV camera-to-world
matrices with the first frame anchored as the origin.  During model forward
passes, intrinsics are normalized by image width and camera translations are
centered and scaled per clip.

\section{Reproducibility, Compute, and Release Details}
\label{app:reproducibility}

\subsection{Training Configuration}

The main submission experiments fine-tune a 1.3B autoregressive video diffusion
transformer initialized from the InSpatio-World/Wan2.1 1.3B checkpoint family
(\citealp{wan2025,inspatio2026world}). Its full-data run uses AdamW at
$10^{-5}$, 100 warmup steps, 30,000 optimization steps, bfloat16, FSDP,
gradient checkpointing, seed 42, and 64 A100-80GB GPUs. The matched recovery,
duration, anchor, and detailed VBench controls in
Appendix~\ref{app:additional-results} instead use the 5B Wan2.2-TI2V backbone
and are labeled separately; their matched training runs use AdamW at
$5\times10^{-6}$ with $\beta=(0.9,0.95)$, weight decay $10^{-4}$, gradient
clipping at 1.0, and 6,000 low-resolution steps on 64 A100 GPUs. In both model
families the VAE and text encoder are frozen, all DiT and PM-RoPE parameters are
trainable, and only newly added PM-RoPE modules are zero-initialized.

For the 1.3B full-data run, 43\% of steps use the V2V family. The remaining
steps sample all-history/node-drop/noisy-memory with relative probabilities
0.56/0.30/0.14; within V2V, 22\% use reference-cache and 25\% of the remainder
use a two-chunk frontier. The resulting total mixture is approximately
31.9\% all-history, 17.1\% node-drop, 8.0\% noisy-memory, 25.2\% plain V2V,
8.4\% frontier, and 9.5\% reference-cache. Reference-cache uses gaps of 12--25
chunks. In contrast, the 5B checkpoint used by the gap-attention probe in
Table~\ref{tab:gap-attention} had reference-cache probability zero; that table
is therefore explicitly a routing diagnostic, not evidence that this 5B
checkpoint learned elapsed-time extrapolation from reference-cache training.

Algorithm~\ref{alg:training-step} specifies one complete training step. Our
scheduler is indexed so that $k=0$ has $\sigma\approx1$ (pure noise) and $k=999$
has $\sigma\approx0$ (clean). An 81-frame video becomes 21 VAE latent frames and
$N=7$ chunks of $m=3$ latent frames. The per-frame loss mask $M$ may contain zero
or soft weights: it changes supervision only. All chunks participate in the same
chunk-causal forward pass, and history K/V tensors are never detached. Therefore,
loss on a recovery frontier backpropagates through the history representations it
reads, even when those history frames have zero direct loss.

\begin{algorithm}[t]
\caption{One \ourmethod{} training step.}
\label{alg:training-step}
\scriptsize
\begin{algorithmic}[1]
\Require video $V$, caption $y$, cameras $C$
\Require event boundaries $E$, degradation peaks $P$, optional reference $V^{\mathrm{ref}}$
\State $x_0\gets\operatorname{VAE}(V)$ \Comment{VAE frozen}
\State $e_y\gets\operatorname{TextEnc}(y)$ \Comment{text encoder frozen}
\State If present, $x_0^{\mathrm{ref}}\gets\operatorname{VAE}(V^{\mathrm{ref}})$
\State Split $x_0$ into $N$ chunks of $m=3$ latent frames; sample $\epsilon\sim\mathcal N(0,I)$
\State Sample $k_c\sim\mathcal U\{0,\ldots,999\}$ for every chunk $c$; set $M\gets\mathbf 1$ and positions $p_i\gets i$
\State Sample regime $r$ from the configured curriculum distribution
\If{$r=\mathrm{NoisyMemory}$}
  \State $k_{1:N-1}\sim\mathcal U\{500,\ldots,984\}$; $M_{1:N-1}\gets0$ \Comment{last chunk is the target}
\ElsIf{$r=\mathrm{NodeDrop}$}
  \For{each segment $S=[E_j,E_{j+1})$}
    \State Drop each $c\in S$ independently with probability $0.35$; if all are dropped, retain one uniformly
    \State $k_c\gets0$ for each dropped $c$ \Comment{pure-noise history; full-clip loss remains}
  \EndFor
\ElsIf{$r\in\{\mathrm{V2VSuffix},\mathrm{V2VFrontier}\}$}
  \State Choose split $s$ from $P$ and pin $k_{c<s}\gets999$ \Comment{clean teacher-forced history}
  \State For suffix, set $M_c=\mathbf 1[c\ge s]$; for frontier $[f,f+w)$, set $M_c=\mathbf 1[f\le c<f+w]$
\ElsIf{$r=\mathrm{ReferenceCache}$}
  \State Prepend clean chunks from $x_0^{\mathrm{ref}}$ and retain the target near its degradation peak
  \State Sample $G\sim\mathcal U\{12,\ldots,25\}$; assign reference positions from 0 and target positions from $Gm$
  \State Pin reference and target seed clean; sample frontier $[f,f+w)$ and set $M_c=\mathbf 1[f\le c<f+w]$
\EndIf
\If{the batch uses I2V-aligned seeding}
  \State Pin the configured first latent frame(s) clean and set their direct loss weights to zero
\EndIf
\State Apply configured pre-peak and recovery-region soft weights to $M$ when applicable
\State $\sigma\gets\Sigma(k)$; $x_k\gets(1-\sigma)x_0+\sigma\epsilon$; $u\gets\epsilon-x_0$
\State $\hat u\gets f_\theta(x_k,k,p,C,e_y;\text{chunk-causal PM-RoPE attention})$
\State $\mathcal L_{\mathrm{flow}}\gets\frac{\sum_i M_i\lVert\hat u_i-u_i\rVert_2^2}{\sum_i M_i\,d}$; $\hat x_0\gets x_k-\sigma\hat u$
\State $\mathcal L\gets\mathcal L_{\mathrm{flow}}+\lambda_{\mathrm{adapt}}\mathcal L_\Delta(\hat x_0,x_0,M)$ \Comment{Appendix~\ref{app:dynamic-loss}}
\State Backpropagate $\mathcal L$ through the single full-sequence graph; clip gradient norm and take one AdamW step
\end{algorithmic}
\end{algorithm}

The main data mixture is the normalized 81-frame format described in
Appendix~\ref{app:data}.  Training sources include OpenVid~\citep{nan2024openvid}, DL3DV~\citep{ling2023dl3dv}, Pexels-based
dynamic clips, Pexels pan/evolution camera-loop data, Veo3 full-frame occlusion
clips, and Helios-generated hard dynamics~\citep{yuan2026helios}.  In the production full-data
configurations, the sampler mixes real DA3-packed~\citep{lin2025depth} sources with generated dynamics
and loop sources.  V2V-heavy runs use degradation-aware recovery training with
frontier continuation and reference-cache reads, while I2V-aligned runs use clean
first-frame seeding and node-drop training to align train/test cache structure.

\subsection{Evaluation Protocols}

STEVO-Bench~\citep{ma2026outofsight} and VBench~\citep{huang2024vbench} I2V
generation follow their respective evaluator snapshots. Our STEVO-Bench
snapshot uses 81 frames at 16 fps
and $480\times832$; each VBench comparison uses a common format across its
reported arms. We use fixed generation seeds and per-sample manifests so
aggregate scores can be traced back to generated videos, prompts, checkpoint
identifiers, and metric outputs.
For ablation tables scored by Qwen3-VL~\citep{bai2025qwen3vl}, we report means over the fixed evaluation
set and release the per-example scores and bootstrap resampling script used to
estimate confidence intervals over examples.

\begin{table}[H]
\centering
\small
\setlength{\tabcolsep}{4pt}
\caption{Protocol and model-size ledger for the quantitative result families.
``Official'' denotes the STEVO-Bench leaderboard evaluator; the other rows use
the controlled protocols defined in this paper.}
\label{tab:protocol-ledger}
\begin{tabular}{p{0.22\linewidth}p{0.19\linewidth}p{0.49\linewidth}}
\toprule
Result family & Backbone & Evaluation provenance \\
\midrule
STEVO-Bench main & 1.3B & Full-data \texttt{checkpoint-1000}; official task list and three-vote aggregates; seed 1000; 25 steps; 81 frames at 16 fps and $480\times832$ \\
VBench I2V & 1.3B baseline / 5B trained arms & Released InSpatio-World baseline and matched 5B data-only/full checkpoints under one runner, seed set, and resolution \\
PM-RoPE, loss, and data ablations & 1.3B & Fixed 200-clip recovery set; seed 42; each Qwen3-VL score averaged over three trials \\
Matched recovery controls & 5B & Matched initialization and 6,000-step budget; 24 fixed-seed cases at $256\times448$ \\
Production fog recovery & 5B & Production checkpoints; 25 denoising steps; 24 held-out, time-aligned cases \\
Efficiency & 1.3B / 5B & DiT-only streaming benchmark on one A100-80GB; median of five runs \\
\bottomrule
\end{tabular}
\end{table}

\subsection{Artifact Map and Causal Probe Entry Points}

Our evaluation codebase keeps probe entry points separate from training. The
held-out fog-transfer experiment in Table~\ref{tab:production-v2v} is
implemented by
\path{tools/run_remind_fog_holdout.py}. It applies a ramp-in/hold/ramp-out fog
event to latent-history chunks 2--4, pins the first five chunks, and scores the
two generated recovery chunks against the time-aligned clean suffix. The
anchor intervention in Table~\ref{tab:anchor-knockout} is implemented by
\path{tools/run_remind_anchor_knockout.py}; it holds the input latents, RoPE
positions, prompt, denoising schedule, and seed fixed while selecting
\texttt{full}, \texttt{no\_anchor}, or \texttt{anchor\_only} cache content.
Both scripts expose checkpoint paths, denoising steps, shard index/count, and
noise seed as command-line arguments, consume the state-v2 task and
source manifests, and write per-case metrics together with generated videos.
Their default evaluation format is 81 frames at 16 fps and
$480\times832$, with seed 20260711. A minimal invocation requires only
\texttt{--output-dir}; raw-base and teacher controls are selected with
\texttt{--skip-generator-checkpoint-load} and \texttt{--no-lora --steps 25},
respectively.

The remaining controlled analyses use the same interface and manifest family:
\path{tools/run_remind_dark_duration.py} produces the duration-conditioned
state-advance measurements in Table~\ref{tab:duration-objective};
\path{tools/run_remind_gap_probe.py} sweeps the hidden RoPE gap and records
per-layer anchor-attention mass for Table~\ref{tab:gap-attention}; and
\path{tools/bench_pmrope_efficiency.py} benchmarks the standard, PM-RoPE, and
dual-attention modes in Table~\ref{tab:efficiency}. The gap runner additionally
writes one generated recovery video and one metrics record per case and gap,
so the archival output package is designed to permit re-judging without
rerunning generation.
Each causal runner writes generated videos and a \path{metrics.json} under each
case directory, plus \path{summary_shard0.json}; shard outputs can be merged by
case identifier without averaging shard means. The efficiency runner writes
one JSON record per attention mode, including per-chunk time, peak allocated
memory, KV-cache size, and analytic final-chunk attention FLOPs. These probe
scripts and their pinned manifests are provenance identifiers here, not a claim
that every entry point is already in the current public repository; they will
be packaged with the archival artifacts.
Given the official per-case STEVO summary and a directory containing the
submitted videos, \path{supplementary/build_stevo_selected_gallery.py}
regenerates Fig.~\ref{fig:stevo-selected-gallery}; the script validates the
joint official judgments and records the legibility-selected task identifiers.

\subsection{Compute Resources}

All reported training and evaluation jobs were run on Lilypad using A100 80GB GPU
workers.  Debug and evaluation jobs used 8 GPUs, the main I2V-aligned and V2V
ablation runs used 32 GPUs, and the full-data production runs used 64 GPUs.
Helios data generation jobs used 8-GPU shards for the curriculum sources used in
the training mix.  We will release the corresponding run configuration records
and per-run GPU-hour accounting so the compute budget can be audited for each
reported table.

\subsection{Public Research Release and License}

The current project page provides released 5B models, DMD-EMA weights,
inference code, and examples. A separate manifest-first data release contains
1,419,331 released rows (1,027,329 Helios generated rows, 265,945 pair-fixed
InSpatio camera-augmentation rows, and 126,057 Kubric camera-augmentation
rows). These are release-manifest rows, not unique-video counts or the realized
exposure census of the historical 1.3B run. The archival research package will additionally
include the training, data-construction, and evaluation code; exact run
configurations; evaluation manifests; prompts and seeds; generated-video
metadata; per-example metric records; and the scripts used for
STEVO-Bench~\citep{ma2026outofsight}, VBench~\citep{huang2024vbench}, causal
probes, and ablation tables. We will include checksums and a
table-to-artifact index so every reported aggregate can be traced to its cases
and checkpoint. The package will use a non-commercial research license rather
than an OSI-approved open-source license. Existing assets retain their upstream
licenses and terms. Public or third-party media that cannot be redistributed
directly will be represented by source identifiers, manifests, derived
annotations, camera/depth metadata, and preprocessing instructions rather than
repackaged raw videos.


\end{document}